\ifdefined\XeTeXversion
\else
  \pdfoutput=1
\fi
\PassOptionsToPackage{table}{xcolor}
\documentclass[10pt,logo,onecolumn,copyright]{nv}

\usepackage{graphicx}
\usepackage{xcolor}
\usepackage[utf8]{inputenc}
\usepackage[T1]{fontenc}
\usepackage{times}
\usepackage{amsmath}
\usepackage{amssymb}
\usepackage{mathtools}
\usepackage{booktabs}
\usepackage{multirow}
\usepackage{tabularx}
\usepackage{float}
\usepackage{url}
\usepackage[square,sort,comma,numbers]{natbib}
\usepackage{xspace}
\usepackage{longtable}
\usepackage{listings}

\usepackage{amsmath,amsfonts,bm}

\def\eqref#1{equation~\ref{#1}}

\def\1{\bm{1}}

\DeclareMathAlphabet{\mathsfit}{\encodingdefault}{\sfdefault}{m}{sl}
\SetMathAlphabet{\mathsfit}{bold}{\encodingdefault}{\sfdefault}{bx}{n}

\definecolor{nvidiagreen}{HTML}{76B900}
\hypersetup{
  colorlinks=true,
  citecolor=nvidiagreen,
  linkcolor=nvidiagreen,
  urlcolor=nvidiagreen
}
\renewcommand{\today}{2026-10-01}

\definecolor{oursgray}{gray}{0.93}
\definecolor{dropcolor}{RGB}{180,65,35}
\definecolor{sdcolor}{RGB}{35,105,145}
\lstdefinestyle{paperprompt}{basicstyle=\fontsize{4}{5}\selectfont\ttfamily,
  columns=fullflexible,breaklines=true,breakatwhitespace=true,
  keepspaces=true,showstringspaces=false,frame=single,framesep=5pt,
  xleftmargin=6pt,xrightmargin=6pt,aboveskip=8pt,belowskip=8pt,
  literate={—}{{\textemdash}}1 {–}{{\textendash}}1}
\makeatletter
\newcommand{\teasercaption}{\def\@captype{figure}\caption}
\makeatother
\newsavebox{\papertablebox}

\graphicspath{{figures/}}

\title{USDCraft: Geometrically Grounded Programmatic Modeling of Articulated 3D Assets for Simulation}

\author{
\parbox{\linewidth}{
\centering
\vspace{-5pt}
{\fontsize{9pt}{16pt}\selectfont\textbf{Chuanrui Zhang\textsuperscript{1\thinspace2\thinspace*}, ~ Zaijia Yang\textsuperscript{2}, ~ Duomin Wang\textsuperscript{1\thinspace\textdagger}, ~ Lu Shi\textsuperscript{1}, ~ Daquan Zhou\textsuperscript{3}, ~ Ruihua Zhang\textsuperscript{1}, ~ Ziwei Wang\textsuperscript{2}}}
\\
\vspace{2.2mm}
{\fontsize{8.7pt}{11pt}\selectfont \textsuperscript{1}NVIDIA \quad \textsuperscript{2}NTU \quad \textsuperscript{3}PKU}
\\
\vspace{0.2em}
{\fontsize{8.2pt}{10pt}\selectfont \textsuperscript{*} This work was conducted during an internship at NVIDIA. \quad \textsuperscript{\textdagger} Project Leader.}
\\
\vspace{5pt}
{\fontsize{9pt}{11pt}\selectfont
\href{https://xingyoujun.github.io/usdcraft}{\raisebox{-0.15em}{\includegraphics[height=1em]{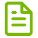}}\hspace{0.35em}\textbf{Project Page}}}
\\
\vspace{2pt}
}
}

\begin{abstract}
Geometrically faithful and functional articulated 3D assets are essential for real-to-sim robot manipulation, where policies trained in simulation must transfer to physical objects.
Recent mesh-based methods learn to infer articulation from annotated 3D assets, but deployment remains challenging when real-world objects fall outside the training distribution or their meshes are incomplete or corrupted.
To address these limitations, we formulate articulated asset reconstruction as programmatic modeling grounded in partial geometric evidence and introduce USDCraft, a framework in which a pretrained LLM writes and revises executable programs for simulation-ready articulated assets without task-specific training.
We propose source geometry analysis, which converts the source mesh into a metric textual description that distinguishes observed surface from unknown space, and iterative geometric rechecking, which re-encodes each candidate in the same representation so that discrepancies point to program edits while unobserved regions remain open to completion.
Visual feedback and physical authoring guidance complete the modeling process, which produces articulated USD assets with explicit physical properties that load into Isaac Sim without manual adjustment.
Experiments demonstrate leading articulation recovery on two benchmarks and validate USDCraft's effectiveness for real-to-sim-to-real robot manipulation.

\end{abstract}

\begin{document}
\maketitle

\vspace{6pt}
\vspace{-0.25cm}
\noindent\begin{minipage}{\linewidth}
\centering
\setlength{\abovecaptionskip}{3pt}
\includegraphics[width=\linewidth]{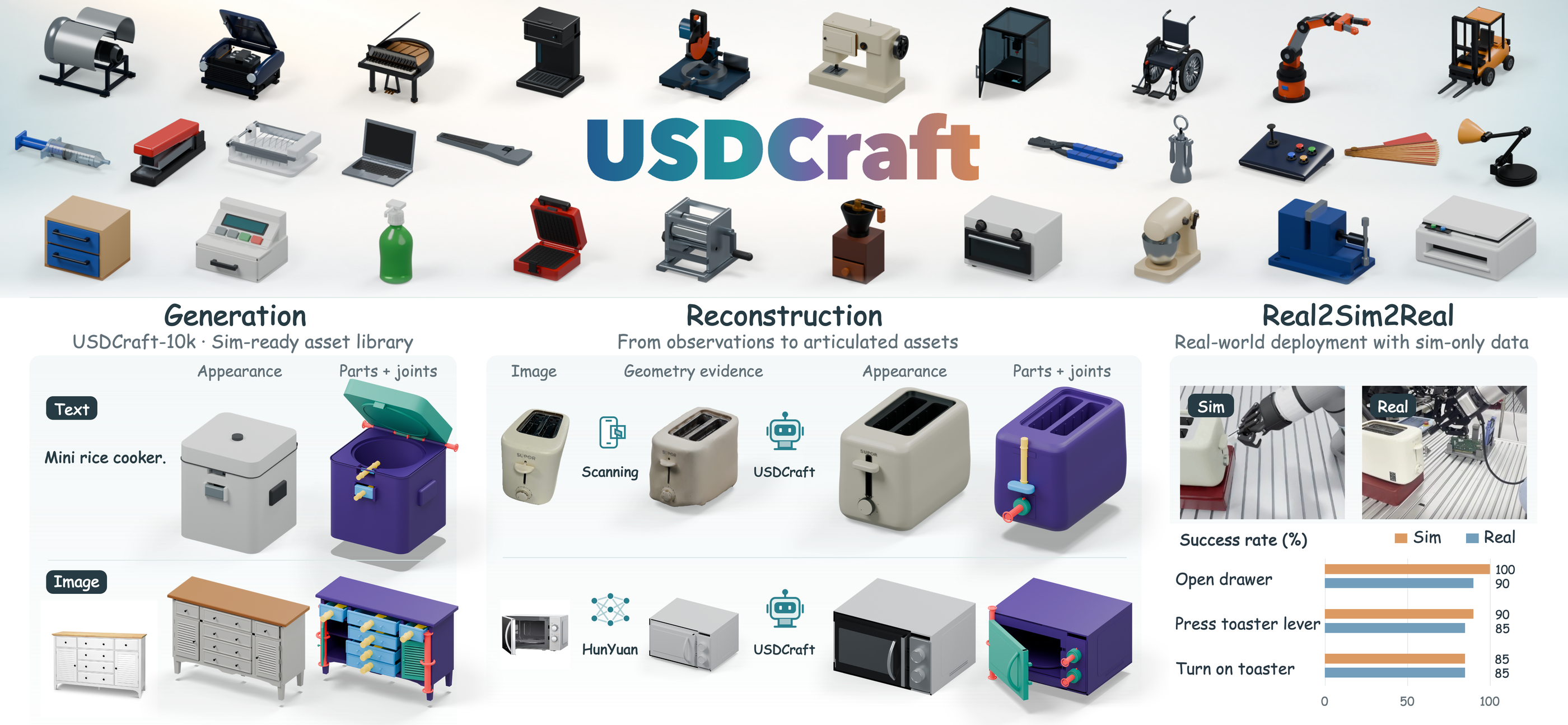}
\teasercaption{USDCraft enables generation, reconstruction, and real-to-sim-to-real manipulation.}
\label{fig:teaser}
\end{minipage}
\vspace{-0.25cm}

\section{Introduction}
\label{sec:intro}

Articulated real-to-sim modeling is attracting growing interest as robot learning increasingly relies on simulated interaction with everyday objects~\citep{zhou2026articraft,cao2025physxanything}.
Demonstration collection and policy training require interactive assets with coherent geometry, functional articulation, and explicit physical properties~\citep{gu2023maniskill2,nasiriany2024robocasa}.
Meeting these needs calls for two complementary workflows for creating simulation-ready assets: generation to supply diverse training objects, and reconstruction to reproduce the geometry and articulation of specific physical objects for simulation-to-real transfer.

Recent mesh-based methods recover articulated objects by partitioning a static mesh into functional parts and estimating the joint axes and connectivity that govern their motion~\citep{li2025particulate,li2026instructparticulate,zhang2026simart}.
These methods face two challenges in real-world deployment.
First, incomplete surfaces and fused components can undermine part segmentation, while partitioning alone cannot recover missing geometry or functional openings.
Second, their reliance on task-specific training makes it difficult to generalize to out-of-distribution objects whose categories or mechanisms are poorly represented in the training data.
Agentic methods build objects from scratch as executable programs, using LLM knowledge of everyday objects to cover diverse categories without requiring a complete input mesh~\citep{zhou2026articraft,lin2026procedura}.
Without measurements of the real object's geometry to guide construction, however, the generated dimensions and part placement may not match the target instance.
Even when these assets run in a simulator, such geometric differences can limit the transfer of learned manipulation policies to the real object.
The challenge is to make reliable source geometry guide both program construction and revision while retaining the flexibility to rebuild missing geometry and separate fused parts.

To address these limitations, we formulate articulated asset reconstruction as programmatic modeling grounded in partial geometric evidence and introduce \textbf{USDCraft}, a framework that realizes this formulation for articulated 3D assets in simulation (Figure~\ref{fig:teaser}).
A pretrained LLM, without task-specific fine-tuning, writes an editable asset program from a source mesh and a reference image, compiles it, and revises it against the source, deciding for itself which evidence to gather and which feedback to consult.
Two components ground this process in measured geometry.
Source geometry analysis converts the mesh into a metric textual description of observed exterior and interior surfaces and lets the agent query further measurements on demand.
Iterative geometric rechecking re-encodes each candidate in the same representation, so that discrepancies with the source point to specific program edits while unobserved regions remain open to justified completion.
Visual feedback from rendered candidates complements these measurements by revealing missing parts, implausible assemblies, and clearance problems across articulated configurations.
The same program representation supports generation from text or images, and every compiled asset carries collision geometry, masses, contact parameters, and joint limits for direct use in Isaac Sim.

Evaluation covers diverse generated objects and real scans in USDCraft-bench and the public Lightwheel dataset~\citep{lightwheel2025simready,li2025particulate}.
USDCraft leads the evaluated baselines on USDCraft-bench and achieves competitive geometry and strong part recovery and joint estimation relative to published Lightwheel results.
Downstream tests validate real-to-sim-to-real manipulation.
We also introduce USDCraft-10k, a library of articulated assets generated with USDCraft.

Our main contributions are:
\begin{itemize}
    \item A formulation of articulated asset reconstruction as programmatic modeling grounded in partial geometric evidence, solved by pretrained LLMs without task-specific training.
    \item Source geometry analysis and iterative geometric rechecking, which make source geometry readable to LLMs and return candidate discrepancies in the same representation.
    \item Empirical validation of reconstruction accuracy and consistency across diverse assets, together with downstream demonstrations of real-to-sim-to-real robot manipulation.
\end{itemize}

\section{Related Work}
\label{sec:related-work}

\paragraph{Mesh-based articulation and learned 3D representations.}
PartNet, AKB-48, and GAPartNet provide complementary resources for hierarchical parts, real-world articulated objects, and transferable part-level perception and manipulation~\citep{mo2019partnet,liu2022akb48,geng2023gapartnet}.
A complementary line strengthens part segmentation through pretrained visual knowledge and learned 3D features, exemplified by PartSLIP, SAMPart3D, and PartField~\citep{liu2023partslip,yang2024sampart3d,liu2025partfield}.
Building on part-level representations, Particulate jointly estimates parts, connectivity, and motion constraints from a static mesh; Instruct-Particulate adds kinematic instructions and heterogeneous supervision to improve cross-category generalization~\citep{li2025particulate,li2026instructparticulate}.
3D language models offer a token-based formulation: SIMART jointly infers part decomposition and kinematics from sparse 3D tokens, while ArtLLM generates articulation-aware layouts that guide part geometry synthesis~\citep{zhang2026simart,wang2026artllm,mandi2024real2code}.
Articulate AnyMesh instead uses pretrained vision-language models and visual prompting without articulation-specific training~\citep{qiu2025anymesh}.
USDCraft follows a programmatic rebuilding route that combines pretrained LLMs' broad object knowledge with measured geometric feedback and iterative correction, supporting accurate reconstruction from imperfect meshes across diverse everyday categories without requiring task-specific articulation training for new categories.

\paragraph{Asset generation and agentic modeling.}
PhysX-Anything learns to generate geometry, articulation, and physical attributes from a single image, while URDF-Anything+ jointly generates part geometry and joints in an autoregressive diffusion model~\citep{cao2025physxanything,wu2026urdfanythingplus}.
Earlier real-to-sim pipelines such as URDFormer and Articulate-Anything predict kinematic structure from images and assemble objects from part meshes drawn from existing asset libraries~\citep{chen2024urdformer,le2024articulateanything}.
Because their geometry comes from the library instead of the target instance, they address a different setting from instance-level reconstruction, and we do not include them as baselines.
Articraft makes programmatic asset generation scalable through a dedicated SDK and a restricted authoring harness that returns structured execution and validation feedback~\citep{zhou2026articraft}.
Procedura uses a more structured assembly process: typed mates constrain part placement, and a separate visual critic guides localized program revisions~\citep{lin2026procedura}.
For real-to-sim reconstruction, however, validating a generated assembly does not establish agreement with the target object's measured geometry.
USDCraft combines source geometry analysis, iterative geometric rechecking, and visual feedback to ground reconstruction in the measured source mesh.

\section{Method}
\label{sec:method}

\begin{figure}[t]
    \centering
    \setlength{\abovecaptionskip}{3pt}
    \includegraphics[width=\linewidth]{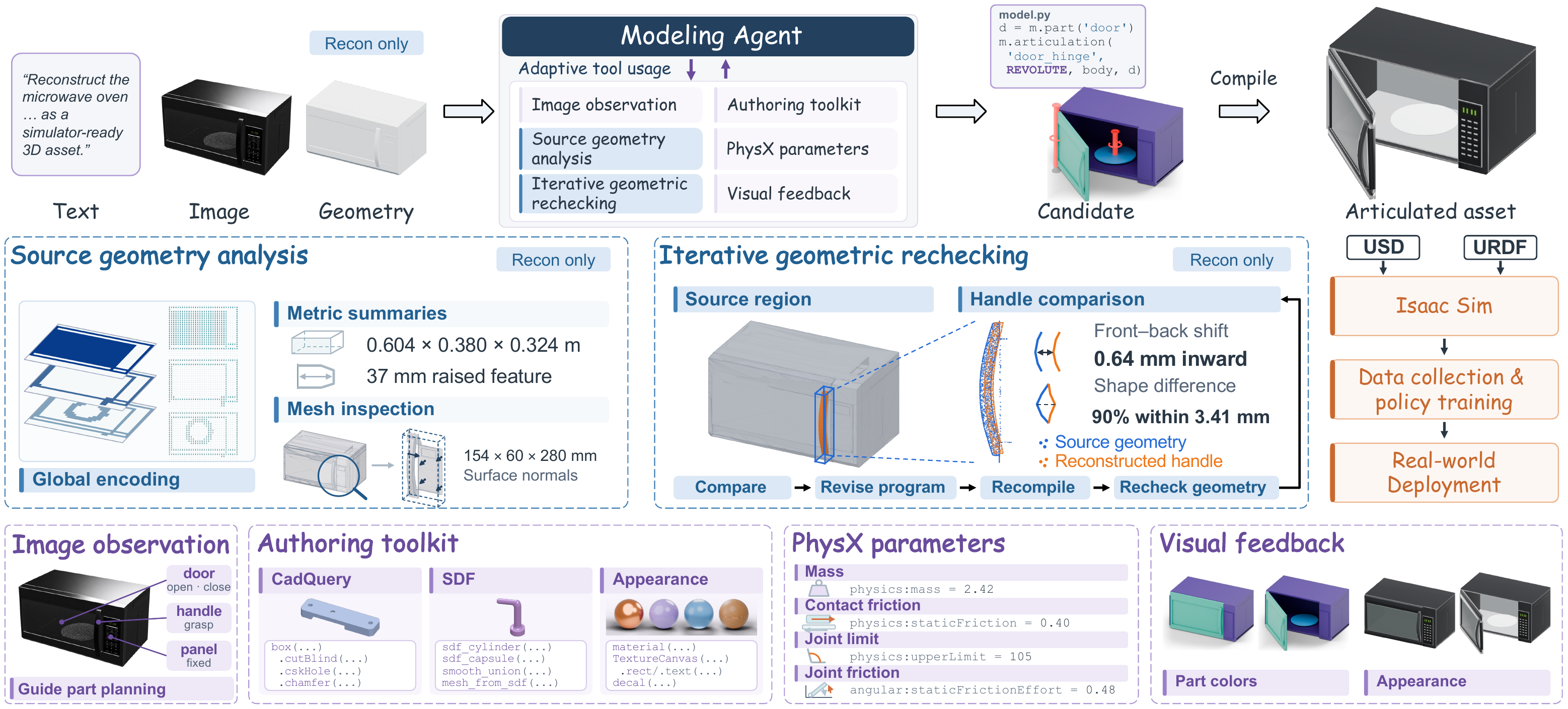}
    \caption{USDCraft overview. The modeling agent refines an asset program, choosing which tools to use and in what order. Blue modules measure the source geometry (reconstruction only), purple modules are shared tools, and orange marks downstream use.}
    \label{fig:overview}
    \vspace{-8pt}
\end{figure}

USDCraft treats articulated asset reconstruction as programmatic modeling grounded in partial geometric evidence (Figure~\ref{fig:overview}).
The modeling agent is an off-the-shelf LLM that iteratively refines an executable asset program against the source, supported by a modeling harness of tools, guidance, and feedback.
Because LLMs cannot read a mesh directly, source geometry analysis converts the source mesh into a metric text representation that separates observed surface from unknown space.
Iterative geometric rechecking compares each candidate with the source in this representation to locate errors for revision, and visual feedback checks appearance and articulated function.

\suppressfloats[t]

\subsection{Problem setup}
\label{sec:method-inputs}

For reconstruction, the input is a source mesh, a reference image, and an optional text request.
The output is an executable asset program that compiles into an articulated asset with rigid parts, joints defined by type, axis, origin, and limits, and physical parameters such as masses and contact friction.
The asset should match the mesh wherever the mesh reliably observes the object, complete missing geometry consistently with the image, and move the way the depicted object works.
The mesh therefore determines dimensions and placement, while the image establishes object identity, visible parts, and likely motion, including components absent from an incomplete scan.
Through image observation, the agent records these parts, how they attach, and how they move as a part plan that guides construction.
Generation is the special case without a source mesh: it skips the mesh-processing tools and conditions the program on a text request, an image, or both.

\subsection{Source geometry analysis}
\label{sec:method-geometry}

Pretrained LLMs reason well over text but cannot read a mesh directly.
Rendered views discard metric scale and hide interior structure, and raw vertex or point lists exceed a practical context budget.
We therefore introduce source geometry analysis, which converts the source mesh into a compact metric description that the agent can read and query.
It combines a global axial geometry encoding with targeted mesh inspection.
The axial encoding samples the surface support of the source mesh on a regular grid in a shared metric frame.
Let $S_M$ denote the set of grid cells that contain sampled source surface.
We serialize $S_M$ as an ordered sequence of axial slice maps along the vertical axis, together with local metric summaries of bounds, protrusions, openings, and observed interior fittings.
Every cell outside $S_M$ is marked as unknown instead of empty or solid.
This distinction matters for incomplete meshes.
A hole in a scan does not imply free space, and the inside of a closed shell does not imply solid material, so the agent remains free to complete such regions from the image and the object's function.
Automatically detected features serve as measurement anchors, and their functional interpretation is left to the agent.
Mesh inspection complements this limited-resolution overview on demand.
Guided by the reference image and its current part plan, the agent renders the source from chosen viewpoints and queries local bounds, point distributions, and surface orientations within selected regions.
Together, the two forms of evidence let an LLM reason about dimensions, part placement, and interior structure without training on 3D mesh understanding.

\subsection{Programmatic asset representation}
\label{sec:method-program}

We represent each asset as an executable program because a program exposes the quantities that later feedback needs to change.
Parts, joints, and physical properties are defined by explicit parameters in the program, so a discrepancy found during inspection can be addressed by a local edit instead of regenerating the asset.
A program can also introduce parts and connections that the source never observed, which segmenting the given surface cannot do.
The authoring toolkit combines parametric solid modeling with CadQuery~\citep{cadquery2026} for dimension-controlled mechanical parts and signed distance field (SDF;~\citealp{curless1996volumetric}) modeling with analytic primitives for organic shapes and smooth transitions.
Appearance is authored in the same program with physically based materials, textures, and decals.
Physical properties are authored directly as PhysX parameters, the quantities used by NVIDIA PhysX, the physics engine of Isaac Sim~\citep{makoviychuk2021isaac}.
The program sets masses, contact friction, joint limits, joint friction, and drives in the engine's own terms, which removes the conversion step needed when physical properties are predicted in a generic form.
These values follow authoring guidance distilled from Isaac Sim test outcomes, which also keeps hand-operated mechanisms passive and checks articulated clearances along normal-use sequences.
Properties that cannot be measured from the evidence are treated as modeling assumptions.
Compiling the program derives collision geometry and exports a USD asset that loads into Isaac Sim with plausible physical behavior and without manual adjustment.

\subsection{Iterative geometric rechecking}
\label{sec:method-rechecking}

Source geometry analysis informs the initial construction, but errors introduced while writing the program must also be found and corrected.
We express this feedback in the same representation as the source evidence.
At revision step $t$, we re-encode the current candidate in its reference configuration on the source grid. 
With $S_t$ denoting the cells that contain sampled candidate surface, the comparison returns source coverage and two discrepancy sets:
\begin{equation}
    r_t = \frac{|S_M\cap S_t|}{|S_M|},
    \qquad \mathcal{D}_t^- = S_M\setminus S_t,
    \qquad \mathcal{D}_t^+ = S_t\setminus S_M.
    \label{eq:recheck-coverage}
\end{equation}
Since $S_M$ and $S_t$ share one grid, each discrepancy appears on axial slices the agent has already read and can be traced back to the part of the program that produced it.
Cells in $\mathcal{D}_t^-$ contain observed source surface that the candidate misses, which is direct evidence of an error.
Cells in $\mathcal{D}_t^+$ lie where the source is unknown.
They may be errors, but they may also be necessary completions such as inner walls, image-supported parts, or connections absent from an incomplete scan.
We therefore treat $\mathcal{D}_t^-$ as a correction signal and $\mathcal{D}_t^+$ as an open question that the agent resolves with the image and the object's function, instead of penalizing both uniformly.

Coverage alone cannot tell whether a part is misplaced or wrongly shaped, and the two cases call for different edits.
For finer comparison, first-surface maps from six axis-aligned directions measure face offsets, silhouette differences, and the placement of openings and protrusions.
The agent can also select a source region and a candidate part and compare their first-surface depths along a chosen viewing direction, without ground-truth part segmentation.
Let $u$ index a shared 2D projection grid, let $d_M(u)$ and $d_t(u)$ be the source and candidate depths, and let $\Omega$ contain the locations observed in both.
With the residual $\delta(u)=d_t(u)-d_M(u)$, we decompose the discrepancy into
\begin{equation}
    b = \operatorname{median}_{u\in\Omega}\delta(u),
    \qquad
    e_{\mathrm{profile}} = Q_{0.9}\!\left(
        \{ |\delta(u)-b| : u\in\Omega \}\right).
    \label{eq:local-profile}
\end{equation}
The median offset $b$ measures the front--back shift along the viewing direction, and $e_{\mathrm{profile}}$, the 90th percentile of the residual after removing this offset, measures the shape difference.
For the handle in the overview, $b$ is only 0.64~mm while $e_{\mathrm{profile}}$ is 3.41~mm, so its profile rather than its placement needs revision; a large $b$ with a small $e_{\mathrm{profile}}$ would instead indicate a misplaced part.
After each revision, the agent recompiles the program and rechecks the affected geometry.

\vspace{-2pt}
\subsection{Visual feedback}
\label{sec:method-visual}

Geometric diagnostics compare surfaces and cannot judge whether an asset looks and works like the intended object.
Visual feedback fills this gap by rendering the candidate with its materials and with per-part colors from viewpoints selected by the agent, at the default or specified joint configurations.
The agent compares these images with the reference image or requested description to find missing components, implausible assemblies, incorrect appearance, and clearance problems that appear only when parts move, and revises the program accordingly.
In generation, where no source mesh is available, visual feedback provides the main check on the candidate.

\section{Experiments}
\label{sec:experiments}
\vspace{-6pt}

The experiments evaluate geometric reconstruction and articulation recovery on USDCraft-bench and Lightwheel.
Harness experiments assess the effectiveness of the USDCraft modeling system, including reconstruction quality, consistency, and transfer across models.
Downstream tests examine real-to-sim-to-real robot manipulation, followed by an evaluation of text and image conditioned asset generation.
Component ablations assess individual tools and feedback mechanisms.
 
\vspace{-4pt}
\subsection{Evaluation setup}
\vspace{-4pt}

\paragraph{Benchmark.}
USDCraft-bench contains 60 agent-generated assets, with 20 each from USDCraft, Articraft, and Procedura, and a 40-object extension combining meshes from Hunyuan3D 2.1~\citep{tencent2025hunyuan21}, and real scans.
All inputs are processed into static meshes without part labels, articulation metadata, or authoring programs; reference annotations remain separate from the model inputs.
Lightwheel~\citep{lightwheel2025simready} contains 243 human-modeled articulated objects.

\vspace{-4pt}
\paragraph{Metrics.}
Part recovery is measured by F1, the harmonic mean of precision and recall over matched parts.
For static geometry, generalized intersection over union (gIoU) and mean intersection over union (mIoU) measure part-box overlap, while part Chamfer distance (PC) measures symmetric mean squared nearest-neighbor distances between corresponding part surfaces.
Fully articulated geometry is assessed with gIoU, PC, and whole-object Chamfer distance (OC).
Joint-axis angular error (AE) measures axis-direction disagreement, and location error (LE) measures the distance between revolute axis lines.
Geometric distances are computed in normalized object coordinates.

\subsection{Reconstruction results of USDCraft}
\label{sec:exp-reconstruction}

\begin{table}[!t]
\begingroup
\renewenvironment{table}[1][]{\begingroup}{\par\endgroup}
\begin{table}[tbp]
\centering
\caption{USDCraft-bench reconstruction results. \textbf{Best}/\underline{second best}. ArtLLM$^{\dagger}$: only success.}
\label{tab:bench100}
\vspace{5pt}
\small\setlength{\tabcolsep}{6pt}
\sbox{\papertablebox}{\begin{tabular}{lccccccccc}
\toprule
 & Part match & \multicolumn{3}{c}{Static geometry} & \multicolumn{3}{c}{Articulated geometry} & \multicolumn{2}{c}{Joint axes} \\
\cmidrule(lr){2-2}\cmidrule(lr){3-5}\cmidrule(lr){6-8}\cmidrule(lr){9-10}
Method & F1(\%)$\uparrow$ & gIoU$\uparrow$ & PC$\downarrow$ & mIoU$\uparrow$ & gIoU$\uparrow$ & PC$\downarrow$ & OC$\downarrow$ & AE$^\circ\downarrow$ & LE$\downarrow$ \\
\midrule
ArtLLM$^{\dagger}$ & 19.337 & 0.194 & 0.136 & 0.390 & 0.189 & 0.373 & 0.043 & 61.156 & 0.366 \\
SIMART & 35.957 & 0.282 & 0.124 & 0.483 & 0.273 & 0.216 & 0.020 & 46.656 & 0.262 \\
Particulate & 33.366 & 0.300 & 0.127 & 0.499 & 0.289 & 0.212 & 0.021 & 48.280 & 0.244 \\
Articraft--Astra & 64.041 & 0.204 & 0.098 & 0.422 & 0.196 & 0.173 & 0.019 & 15.106 & 0.092 \\
\rowcolor{oursgray}
USDCraft--Sol & \underline{79.835} & \underline{0.586} & \underline{0.062} & \underline{0.699} & \underline{0.568} & \underline{0.118} & \underline{0.014} & \underline{9.155} & \underline{0.068} \\
\rowcolor{oursgray}
USDCraft--Astra & \textbf{84.830} & \textbf{0.694} & \textbf{0.052} & \textbf{0.784} & \textbf{0.670} & \textbf{0.093} & \textbf{0.010} & \textbf{6.833} & \textbf{0.044} \\
\bottomrule
\end{tabular}}
\typeout{TABLE-WIDTH-bench100=\the\wd\papertablebox; LIMIT=\the\linewidth}
\usebox{\papertablebox}
\end{table}

\par\vspace{4pt}
\begin{table}[tbp]
\centering
\caption{Lightwheel reconstruction across generation, segmentation, and rebuilding. +G: shared object-function guidance. $^{\ddagger}$: prediction and reference independently centered and scale-normalized.}
\label{tab:lightwheel}
\vspace{5pt}
\footnotesize\setlength{\tabcolsep}{1.3pt}
\sbox{\papertablebox}{\begin{tabular}{@{}lccccccccccc@{}}
\toprule
 & \multicolumn{2}{c}{Input} & Part match & \multicolumn{3}{c}{Static geometry} & \multicolumn{3}{c}{Articulated geom.} & \multicolumn{2}{c}{Joint axes} \\
\cmidrule(lr){2-3}\cmidrule(lr){4-4}\cmidrule(lr){5-7}\cmidrule(lr){8-10}\cmidrule(lr){11-12}
Method & Image & Mesh & F1(\%)$\uparrow$ & gIoU$\uparrow$ & PC$\downarrow$ & mIoU$\uparrow$ & gIoU$\uparrow$ & PC$\downarrow$ & OC$\downarrow$ & AE$^\circ\downarrow$ & LE$\downarrow$ \\
\midrule
\multicolumn{12}{@{}l}{\textit{Image-conditioned generation}} \\
PhysX-Anything & $\checkmark$ &  & 19.700 & -0.458 & 0.247 & 0.093 & -0.459 & 0.334 & 0.064 & 38.800 & 0.123 \\
URDF-Anything+ & $\checkmark$ &  & 35.700 & -0.133 & 0.259 & 0.260 & -0.138 & 0.305 & 0.033 & 50.400 & 0.128 \\
\midrule
\multicolumn{12}{@{}l}{\textit{Mesh Segmentation}} \\
Articulate AnyMesh &  & $\checkmark$ & 57.600 & 0.172 & 0.190 & 0.452 & 0.158 & 0.237 & 0.010 & 21.700 & 0.043 \\
Particulate &  & $\checkmark$ & 65.500 & 0.332 & 0.168 & 0.576 & 0.305 & 0.208 & \underline{0.009} & 20.900 & 0.040 \\
Instruct-Particulate &  & HY3D-3.1 & 64.500 & 0.177 & 0.102 & 0.405 & 0.164 & 0.161 & 0.015 & 18.100 & 0.139 \\
Instruct-Particulate &  & Dataset & 83.400 & 0.583 & 0.091 & \textbf{0.724} & 0.542 & 0.108 & \textbf{0.004} & 13.900 & 0.018 \\
\midrule
\multicolumn{12}{@{}l}{\textit{Mesh Rebuild}} \\
Articraft--Astra$^{\ddagger}$ & $\checkmark$ & & 67.037 & -0.091 & 0.109 & 0.264 & -0.097 & 0.141 & 0.012 & \underline{7.657} & 0.042 \\
\rowcolor{oursgray}
USDCraft--Astra & $\checkmark$ & $\checkmark$ & \underline{86.983} & \underline{0.590} & \underline{0.050} & 0.683 & \underline{0.555} & \underline{0.088} & 0.011 & 10.151 & \textbf{0.009} \\
\rowcolor{oursgray}
USDCraft--Astra +G & $\checkmark$ & $\checkmark$ & \textbf{89.765} & \textbf{0.631} & \textbf{0.047} & \underline{0.708} & \textbf{0.595} & \textbf{0.088} & 0.010 & \textbf{4.463} & \underline{0.010} \\
\bottomrule
\end{tabular}}
\typeout{TABLE-WIDTH-lightwheel=\the\wd\papertablebox; LIMIT=\the\linewidth}
\usebox{\papertablebox}
\end{table}

\endgroup
\vspace{-4pt}
\end{table}

\begin{figure}[!t]
  \centering
  \setlength{\abovecaptionskip}{3pt}
  \includegraphics[width=\linewidth]{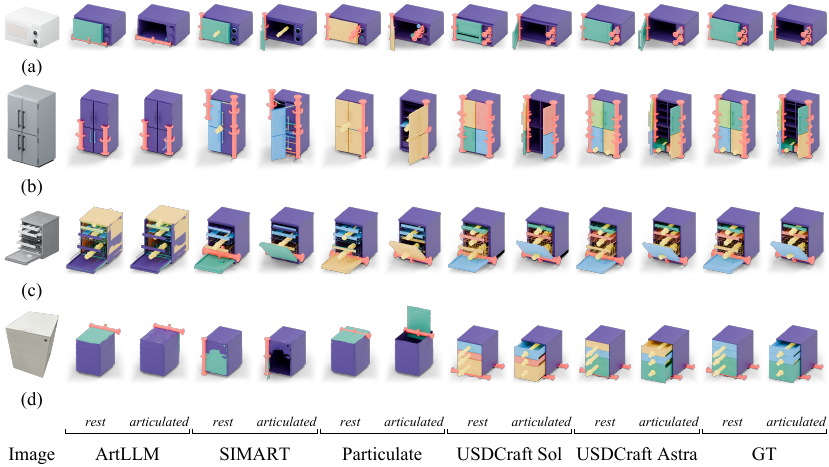}
  \caption{USDCraft-bench reconstructions; each pair shows rest (left) and articulated (right) poses. Red rotation glyphs and yellow double arrows indicate revolute and prismatic joints, respectively.}
  \label{fig:bench-comparison}
  \vspace{-12pt}
\end{figure}

We compare USDCraft's geometric reconstruction and articulation recovery with ArtLLM~\citep{wang2026artllm}, SIMART~\citep{zhang2026simart}, Particulate~\citep{li2025particulate}, and \mbox{Articraft~\citep{zhou2026articraft}} on USDCraft-bench (Table~\ref{tab:bench100}).
ArtLLM scores include only successful outputs, excluding cases with no reconstruction due to P3-SAM segmentation failure.
Both USDCraft--Sol (\mbox{GPT-5.6 Sol}; high reasoning effort, Sol/high) and USDCraft--Astra (\mbox{GPT-6 Astra},~\citealp{openai2026astra}; low reasoning effort, Astra/low) outperform all four baselines on every metric; USDCraft--Astra results are means over three runs, whose variation is far smaller than these margins.
The baselines fail in different ways.
SIMART and Particulate can only partition the given surface, so parts fused in a generated mesh or scan stay merged and parts the mesh misses cannot be added.
Articraft--Astra builds programs from the image alone and recovers more parts with reasonable joint directions, but without measurements its parts are poorly sized and placed.
USDCraft combines the two strengths: rebuilding separates fused parts and completes missing ones, while source geometry analysis fixes their dimensions and positions, improving both geometric accuracy and joint location (Figure~\ref{fig:bench-comparison}).

On Lightwheel's human-modeled assets, we compare USDCraft--Astra with image-conditioned generation, mesh segmentation, and programmatic rebuilding methods (Table~\ref{tab:lightwheel}), using the metrics of Instruct-Particulate~\citep{li2026instructparticulate}.
Without any guidance, USDCraft--Astra achieves the best part matching, static and articulated gIoU, part Chamfer distances, and joint location error, including against Instruct-Particulate applied to the original dataset meshes.
Mesh-segmentation methods keep a lower whole-object Chamfer, and Instruct-Particulate a higher mIoU, as they partition the supplied mesh and reproduce its surface exactly, whereas USDCraft rebuilds every part and accumulates small surface deviations.
Articraft--Astra shows a lower angular error, but joint errors are averaged only over matched parts, and Articraft matches only about two-thirds of the reference movable parts (recall 66.8\% vs.\ 85.0\% for USDCraft), so its errors cover a smaller subset.
The +G setting gives every case the same plain-text description of articulation conventions for all 14 categories, without category labels or instance-specific annotations.
Its further gains show that users can steer how USDCraft defines parts and joints through text, without retraining or changing the method.

\subsection{Effectiveness of the USDCraft harness}
\label{sec:exp-harness}

\begin{table}[!t]
\centering
\caption{Harness comparison with Mini Workflow and USDCraft transfer across models.}
\label{tab:harness100}
\vspace{5pt}
\fontsize{8}{10}\selectfont\setlength{\tabcolsep}{3.5pt}
\sbox{\papertablebox}{\begin{tabular}{lccccccccccc}
\toprule
 & Part match & \multicolumn{3}{c}{Static geometry} & \multicolumn{3}{c}{Articulated geometry} & \multicolumn{2}{c}{Joint axes}  & Time cost & Turns \\
\cmidrule(lr){2-2}\cmidrule(lr){3-5}\cmidrule(lr){6-8}\cmidrule(lr){9-10}\cmidrule(lr){11-11}\cmidrule(lr){12-12}
Method & F1(\%)$\uparrow$ & gIoU$\uparrow$ & PC$\downarrow$ & mIoU$\uparrow$ & gIoU$\uparrow$ & PC$\downarrow$ & OC$\downarrow$ & AE$^\circ\downarrow$ & LE$\downarrow$  & (min) & (\#) \\
\midrule
Mini Workflow--Sol & 73.342 & 0.447 & 0.085 & 0.599 & 0.431 & 0.127 & 0.016 & 24.638 & 0.135 & 5.7 & 24.9 \\
Mini Workflow--Astra & 74.480 & 0.512 & 0.070 & 0.646 & 0.495 & 0.119 & 0.018 & 21.141 & 0.117 & 3.0 & 15.4 \\
\midrule
\rowcolor{oursgray}
USDCraft--Sol & 79.835 & 0.586 & \underline{0.062} & 0.699 & 0.568 & 0.118 & 0.014 & 9.155 & 0.068 & 7.9 & 25.4 \\
\rowcolor{oursgray}
USDCraft--Astra & \textbf{84.830} & \textbf{0.694} & \textbf{0.052} & \textbf{0.784} & \textbf{0.670} & \underline{0.093} & \underline{0.010} & \textbf{6.833} & \underline{0.044} & 4.2 & 16.7 \\
\midrule
\rowcolor{oursgray}
USDCraft--Claude & \underline{82.014} & 0.634 & 0.066 & 0.745 & 0.611 & \textbf{0.090} & \textbf{0.008} & \underline{7.622} & \textbf{0.035} & 27.5 & 59.8 \\
\rowcolor{oursgray}
USDCraft--Gemini & 80.940 & \underline{0.642} & 0.073 & \underline{0.754} & \underline{0.616} & 0.119 & 0.013 & 14.809 & 0.090 & 20.6 & 45.3 \\
\bottomrule
\end{tabular}}
\typeout{TABLE-WIDTH-harness100=\the\wd\papertablebox; LIMIT=\the\linewidth}
\usebox{\papertablebox}
\end{table}

\begin{figure}[!t]
  \centering
  \setlength{\abovecaptionskip}{3pt}
  \includegraphics[width=\linewidth]{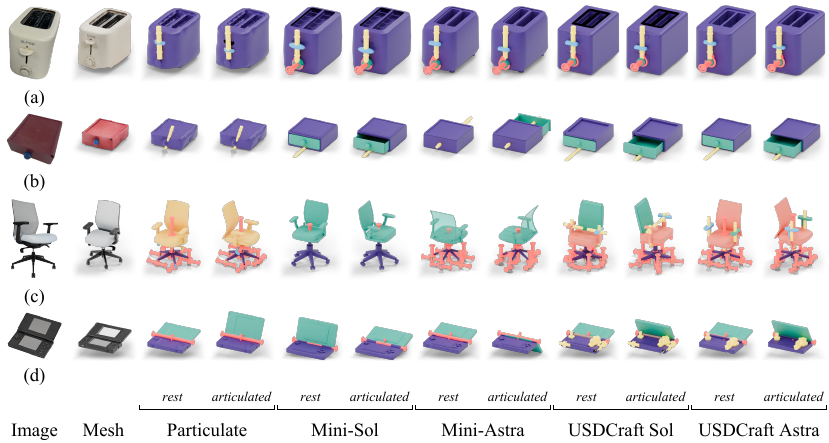}
  \caption{Harness comparison on scanned toaster (a) and drawer (b), and generated chair (c) and game console (d). Pairs show first-run reconstructions at rest (left) and articulated (right) poses.}
  \label{fig:toast-harness}
  \vspace{-12pt}
\end{figure}

To assess the modeling harness, we compare USDCraft with Mini Workflow, which uses the same Sol or Astra backbone and image and mesh inputs but relies on general tools for inspection, construction, and revision (Table~\ref{tab:harness100}).
Its prompt requests the same measured, simulation-ready reconstruction, so the difference lies in the tools and feedback rather than the task.
USDCraft improves all nine reconstruction metrics with both backbones, with the largest gains in the joint axes.
Mini Workflow must write its own measurement code over the raw mesh, whereas source geometry analysis exposes hinge-bearing features such as openings, rails, and handles as metric anchors, and iterative geometric rechecking reveals misplaced parts, so joint axes are anchored to measured geometry.
Tool support matters more than the backbone here, since USDCraft--Sol outperforms Mini Workflow--Astra on every metric (Figure~\ref{fig:toast-harness}).
USDCraft--Claude (Claude Opus 5;~\citealp{anthropic2026claudeopus5}) and USDCraft--Gemini (Gemini 3.8 Flash;~\citealp{google2026gemini38flash}) also outperform Mini Workflow on nearly all metrics, showing that the harness transfers across model families without modification.
The added analysis and rechecking cost about one to two extra minutes per asset for the GPT backbones.

\subsection{Real-to-sim-to-real manipulation}
\label{sec:exp-sim2real}

\begin{table}[tbp]
\centering
\caption{Simulation/real success (\%) with various reconstruction methods. I/M: image/mesh.}
\label{tab:sim2real}
\vspace{5pt}
\small\setlength{\tabcolsep}{3.5pt}
\sbox{\papertablebox}{\begin{tabular}{@{}llcccccc@{}}
\toprule
 & & \multicolumn{2}{c}{Open drawer} & \multicolumn{2}{c}{Press toast} & \multicolumn{2}{c}{Turn on toaster} \\
\cmidrule(lr){3-4}\cmidrule(lr){5-6}\cmidrule(lr){7-8}
Asset method & Input & Sim & Real & Sim & Real & Sim & Real \\
\midrule
Articraft--Sol & I & 90\% & $20\%_{\color{dropcolor}\downarrow70}$ & 90\% & $10\%_{\color{dropcolor}\downarrow80}$ & 80\% & $20\%_{\color{dropcolor}\downarrow60}$ \\
Particulate & M & 0\% & 0\% & 50\% & $10\%_{\color{dropcolor}\downarrow40}$ & 0\% & 0\% \\
Mini Workflow--Sol & I+M & 95\% & $10\%_{\color{dropcolor}\downarrow85}$ & 85\% & $30\%_{\color{dropcolor}\downarrow55}$ & 90\% & $25\%_{\color{dropcolor}\downarrow65}$ \\
Mini Workflow--Astra & I+M & 100\% & $0\%_{\color{dropcolor}\downarrow100}$ & 90\% & $60\%_{\color{dropcolor}\downarrow30}$ & 85\% & $70\%_{\color{dropcolor}\downarrow15}$ \\
\rowcolor{oursgray}
USDCraft--Sol & I+M & 95\% & $85\%_{\color{dropcolor}\downarrow\mathbf{10}}$ & 80\% & $75\%_{\color{dropcolor}\downarrow\mathbf{5}}$ & 85\% & $80\%_{\color{dropcolor}\downarrow5}$ \\
\rowcolor{oursgray}
USDCraft--Astra & I+M & 100\% & $90\%_{\color{dropcolor}\downarrow\mathbf{10}}$ & 90\% & $85\%_{\color{dropcolor}\downarrow\mathbf{5}}$ & 85\% & $85\%_{\color{dropcolor}\downarrow\mathbf{0}}$ \\
\bottomrule
\end{tabular}}
\typeout{TABLE-WIDTH-sim2real=\the\wd\papertablebox; LIMIT=\the\linewidth}
\usebox{\papertablebox}
\vspace{-8pt}
\end{table}

The real-to-sim-to-real study compares policy transfer using assets from USDCraft, Mini Workflow, Particulate~\citep{li2025particulate}, and Articraft~\citep{zhou2026articraft} on drawer opening, toaster-lever pressing, and toaster switching (Table~\ref{tab:sim2real}).
For each method--task setting whose asset supports the required interaction, we collect 200 simulation demonstrations and train Diffusion Policy~\citep{chi2023diffusionpolicy} for 80 epochs with batch size 128, measuring success in each method's asset environment and on the corresponding real object.
Since most assets allow high simulation success, the informative quantity is the sim-to-real drop.
A policy learns where to grasp and push from the asset, so a handle, lever, or switch that is misplaced or missized in simulation sends the real robot to the wrong location.
USDCraft loses at most 10 points on any task, whereas Articraft, built from the image alone, loses 60--80 points, and Mini Workflow varies widely, its Astra variant failing on drawer opening because its misplaced handle causes unsafe contact.
Particulate's incomplete reconstructions support only lever pressing, showing that structural completeness must come before geometric accuracy.

\vspace{-4pt}
\subsection{Text- and image-conditioned generation}
\label{sec:exp-generation}

For generation without source meshes, we compare USDCraft--Sol and USDCraft--Astra against Articraft--Sol~\citep{zhou2026articraft} and Procedura--Sol~\citep{lin2026procedura} on 50 text-conditioned and 50 image-conditioned requests per setting.
An anonymous Sol/high evaluator scores the resulting assets on condition adherence, structural quality, articulation functionality, completeness, and appearance (Table~\ref{tab:generation}).
In this model-based evaluation, USDCraft scores highest in every dimension while taking the least time.
All methods score high on articulation and completeness, and the larger gaps lie in condition adherence and appearance, which USDCraft addresses by authoring materials, textures, and decals in the program and by checking each candidate against the request with visual feedback.
Procedura authors parts through separate agent calls, which makes its runtime grow with the number of parts and may explain its lower structure score, whereas USDCraft writes one program for the whole object.
These results show that the authoring toolkit built for reconstruction also serves generation.
Using this toolkit, we generated USDCraft-10k, a library of 10,000 assets spanning more than 500 everyday object categories, produced entirely with USDCraft--Astra from 70\% text-conditioned and 30\% image-conditioned requests.

\begin{table}[!t]
\begingroup
\renewenvironment{table}[1][]{\begingroup}{\par\endgroup}
\begin{table}[tbp]
\centering
\caption{Generation comparison. Scores are averaged over anonymous evaluations by Sol/high.}
\label{tab:generation}
\vspace{5pt}
\small\setlength{\tabcolsep}{3.5pt}
\sbox{\papertablebox}{\begin{tabular}{@{}lccccccc@{}}
\toprule
 & \multicolumn{5}{c}{Quality score $\uparrow$ (out of 5)} & \multicolumn{2}{c}{Time $\downarrow$ (min)} \\
\cmidrule(lr){2-6}\cmidrule(lr){7-8}
Method & Condition & Structure & Articulation & Complete & Appearance & Mean & Median \\
\midrule
Articraft--Sol & 4.040 & 4.220 & 4.520 & 4.520 & 3.670 & 12.0 & 10.3 \\
Procedura--Sol & 3.810 & 3.380 & 4.590 & 4.620 & 3.440 & 58.8 & 58.7 \\
\rowcolor{oursgray}
USDCraft--Sol & \textbf{4.270} & \textbf{4.270} & 4.720 & \textbf{4.710} & 3.810 & 8.6 & 7.7 \\
\rowcolor{oursgray}
USDCraft--Astra & 4.250 & \textbf{4.270} & \textbf{4.750} & 4.660 & \textbf{3.880} & \textbf{5.6} & \textbf{4.9} \\
\bottomrule
\end{tabular}}
\typeout{TABLE-WIDTH-generation=\the\wd\papertablebox; LIMIT=\the\linewidth}
\usebox{\papertablebox}
\end{table}

\par\vspace{2pt}
\begin{table}[!t]
\centering
\caption{Cumulative ablations on USDCraft-bench (Astra/low). $\hookrightarrow\!+$ adds to the row above.}
\label{tab:ablation-cumulative}
\vspace{5pt}
\footnotesize\setlength{\tabcolsep}{2.7pt}
\sbox{\papertablebox}{\begin{tabular}{lccccccccc}
\toprule
 & Part match & \multicolumn{3}{c}{Static geometry} & \multicolumn{3}{c}{Articulated geometry} & \multicolumn{2}{c}{Joint axes} \\
\cmidrule(lr){2-2}\cmidrule(lr){3-5}\cmidrule(lr){6-8}\cmidrule(lr){9-10}
Setting & F1(\%)$\uparrow$ & gIoU$\uparrow$ & PC$\downarrow$ & mIoU$\uparrow$ & gIoU$\uparrow$ & PC$\downarrow$ & OC$\downarrow$ & AE$^\circ\downarrow$ & LE$\downarrow$ \\
\midrule
Mini Workflow & 74.480 & 0.512 & 0.070 & 0.646 & 0.495 & 0.119 & 0.018 & 21.141 & 0.117 \\
\rowcolor{oursgray}
$\hookrightarrow\!+$ Authoring harness & 80.925 & 0.630 & 0.059 & 0.739 & 0.608 & 0.100 & 0.013 & 9.201 & 0.060 \\
\rowcolor{oursgray}
$\hookrightarrow\!+$ Source geometry analysis & 84.030 & 0.673 & 0.058 & 0.770 & 0.652 & 0.101 & 0.011 & 7.337 & 0.046 \\
\rowcolor{oursgray}
$\hookrightarrow\!+$ Iterative rechecking & \underline{84.293} & \underline{0.689} & \underline{0.054} & \underline{0.783} & \underline{0.662} & \underline{0.096} & \textbf{0.009} & \underline{7.079} & \textbf{0.041} \\
\rowcolor{oursgray}
$\hookrightarrow\!+$ Visual feedback (USDCraft) & \textbf{84.830} & \textbf{0.694} & \textbf{0.052} & \textbf{0.784} & \textbf{0.670} & \textbf{0.093} & \underline{0.010} & \textbf{6.833} & \underline{0.044} \\
\bottomrule
\end{tabular}}
\typeout{TABLE-WIDTH-ablation-cumulative=\the\wd\papertablebox; LIMIT=\the\linewidth}
\usebox{\papertablebox}
\end{table}

\par\vspace{2pt}
\begin{table}[!t]
\centering
\caption{Independent component ablations on USDCraft-bench (Astra/low).}
\label{tab:ablation}
\vspace{5pt}
\footnotesize\setlength{\tabcolsep}{3.0pt}
\sbox{\papertablebox}{\begin{tabular}{lccccccccc}
\toprule
 & Part match & \multicolumn{3}{c}{Static geometry} & \multicolumn{3}{c}{Articulated geometry} & \multicolumn{2}{c}{Joint axes} \\
\cmidrule(lr){2-2}\cmidrule(lr){3-5}\cmidrule(lr){6-8}\cmidrule(lr){9-10}
Setting & F1(\%)$\uparrow$ & gIoU$\uparrow$ & PC$\downarrow$ & mIoU$\uparrow$ & gIoU$\uparrow$ & PC$\downarrow$ & OC$\downarrow$ & AE$^\circ\downarrow$ & LE$\downarrow$ \\
\midrule
\rowcolor{oursgray}
USDCraft & \textbf{84.830} & \textbf{0.694} & \underline{0.052} & \underline{0.784} & \underline{0.670} & \textbf{0.093} & \underline{0.010} & \textbf{6.833} & \textbf{0.044} \\
+ Independent observation agent & 82.556 & 0.664 & 0.059 & 0.765 & 0.640 & 0.103 & 0.011 & 7.849 & 0.055 \\
w/o Source geometry analysis & 81.975 & 0.616 & 0.056 & 0.717 & 0.596 & 0.098 & 0.011 & 10.226 & 0.064 \\
w/o Iterative rechecking & 82.936 & 0.676 & 0.054 & 0.769 & 0.652 & \underline{0.093} & \textbf{0.010} & 7.748 & 0.054 \\
w/o Visual feedback & \underline{83.589} & \underline{0.694} & \textbf{0.050} & \textbf{0.785} & \textbf{0.673} & 0.094 & 0.011 & \underline{7.565} & \underline{0.046} \\
\bottomrule
\end{tabular}}
\typeout{TABLE-WIDTH-ablation=\the\wd\papertablebox; LIMIT=\the\linewidth}
\usebox{\papertablebox}
\end{table}

\endgroup
\vspace{-4pt}
\end{table}

\subsection{Component ablations}
\label{sec:exp-ablation}

Table~\ref{tab:ablation-cumulative} adds components to Mini Workflow one at a time, and Table~\ref{tab:ablation} removes them individually from the complete system.
The authoring harness gives the largest single gain, especially for joint axes: its SDK defines each joint by type, origin, axis, and limits relative to the parts it connects, whereas Mini Workflow must place joint anchors and axes directly in USD coordinates.
Source geometry analysis brings the next largest gain, and removing it causes the largest drop in part recovery, geometric overlap, and joint accuracy, since later rechecking can correct a candidate but cannot supply the measurements needed to plan it.
Iterative rechecking mainly improves geometric fidelity and joint placement, consistent with its role of locating misplaced and misshaped parts.
Removing visual feedback leaves geometry essentially unchanged but lowers part recovery and joint accuracy, indicating that it catches missing parts and implausible motion rather than surface errors.
Adding an independent observation agent lowers every metric, suggesting that an image-only part inventory can mislead the modeling agent, which already performs image observation within the loop.

\section{Conclusion}
\label{sec:conclusion}
USDCraft reconstructs and generates simulation-ready articulated assets by letting a pretrained LLM write, compile, and revise asset programs against measured source geometry.
Source geometry analysis makes the mesh readable to the agent, iterative geometric rechecking turns candidate errors into program edits, and visual feedback catches missing parts and implausible motion.
This grounding improves geometry and articulation over segmentation-based and image-only programmatic methods, matters more than the choice of backbone, and yields assets whose simulation-trained policies transfer to real objects with little loss.
\textbf{Limitations:} fine lattices such as racket strings and cart wire frames remain hard to reconstruct faithfully, and deformable objects are supported only in annotated states and rest geometry, as flexible-body dynamics have not been validated in Isaac Sim.

{
  \small
  \bibliographystyle{unsrtnat}
  \bibliography{references}
}

\clearpage
\section*{Guide to the supplementary material}
The supplementary material provides implementation details, benchmark and evaluation protocols, additional results, and prompts.
\begin{itemize}
    \item Appendix~\ref{sec:additional-method}: harness execution and tool access, the authoring toolkit and compilation, source preparation, and source geometry analysis details.
    \item Appendix~\ref{sec:benchmark-provenance}: input preparation, reference annotation, and object-category coverage.
    \item Appendix~\ref{sec:appendix-protocols}: reconstruction evaluation, part matching and score aggregation, and generation scoring.
    \item Appendix~\ref{sec:additional-experiments}: run-to-run consistency, additional reconstruction comparisons, reconstruction and generation galleries, benchmark-source sensitivity, evaluation in the reference frame, and token usage and cost.
    \item Appendix~\ref{sec:sim2real-protocol}: object scanning and reconstruction, scene assembly, manipulation protocols, and simulation and real-world task sequences.
    \item Appendix~\ref{sec:qualitative-protocol}: fine-lattice reconstruction failures and limitations of deformable-object support.
    \item Appendix~\ref{sec:prompts}: modeling agent instructions, the Mini Workflow prompt, the generation-evaluation rubric, Lightwheel articulation conventions (+G), gallery prompts, and a source geometry analysis example.
\end{itemize}

\clearpage
\appendix
\raggedbottom
\section{Implementation Details}
\label{sec:additional-method}

\subsection{Harness execution and tool access}
The modeling agent starts with the task, instance evidence, PhysX authoring guidance, authoring toolkit foundations, and a documentation inventory.
Detailed API references and worked examples are retrieved as needed, keeping the initial context focused on the modeling task.
A single modeling agent maintains the asset program and selects construction, inspection, and revision actions within the same session.
After an edit, the agent triggers recompilation through a tool; the host compiles the program, refreshes the candidate used for geometric checks and visual feedback, and verifies that the exported asset is complete and can be opened, while the agent interprets discrepancies and decides on repairs.
The instructions for the modeling agent are summarized in Appendix~\ref{sec:author-prompt}.

\subsection{Authoring toolkit and compilation}
\label{sec:toolkit-details}
The modeling agent writes the asset program in \texttt{model.py}.
Worked examples and reusable object models provide starting points that the agent can adapt to the target asset.
SDF shapes are composed with Boolean and smooth-blend operations and meshed at compile time.
Appearance uses physically based rendering (PBR) materials, UV-mapped image textures, and surface decals, with image maps for base color, roughness, metallicity, normals, and occlusion.
The agent can generate texture images procedurally, such as wood grain and fabric patterns, and revise materials and textures through visual feedback.
The PhysX authoring guidance was developed in a separate iterative process that annotated Isaac Sim test outcomes, analyzed failures, and distilled recurring findings into authoring guidelines.
A clean compilation regenerates the asset geometry, articulation, and physical properties, derives collision geometry, and exports USD entries for PhysX and Newton together with a URDF description, carrying materials, masses, contact parameters, and joint settings and packaging referenced meshes and textures with the output.

\subsection{Source preparation and evidence interpretation}
\label{sec:source-preparation}
Source geometry is expressed in a metric, Z-up frame using declared units and axes.
For suspected unit-normalized inputs without a known physical scale, an image-based estimate supplies a scale hypothesis that is then held fixed during modeling; evaluation in the reference frame, which retains errors in this scale, is reported in Appendix~\ref{sec:reference-frame}.
In the ablation of Table~\ref{tab:ablation} only, an independent observation agent supplies an advisory inventory of visible parts and likely motion, which the modeling agent can revise using the original image and mesh.

\subsection{Source geometry analysis details}
\label{sec:axial-encoding-details}
For the axial encoding in source geometry analysis, geometric support is sampled from ray intersections along three coordinate axes, supplemented by mesh vertices and triangle centroids, including observed surfaces inside outer shells.
The fine grid uses 192 cells along the longest extent.
Support is summarized as XY slices ordered along Z, using the finest of 32, 24, or 16 longest-axis cells that fits the text budget, and repeated slices are stored once.
Metric bounds and axis directions connect the text to source coordinates.
Local summaries of bounds, protrusions, openings, and interior fittings retain detail beyond the coarse slices.
The modeling agent receives a compact digest and can inspect the fuller encoding or request local measurements when needed.
Appendix~\ref{sec:axial-example} provides an example of the resulting text encoding.

\section{Benchmarks and Annotation}
\label{sec:benchmark-provenance}

\subsection{Input preparation and reference annotation}
Inputs are converted to a single static mesh with one material: transforms are baked, part names and hierarchy are removed, and joint, animation, and custom metadata are stripped.
UVs and texture atlases are rebuilt, and vertex and face indices are reordered consistently across sources.
The process preserves observed surface geometry and appearance.

Generated references retain their original articulation annotations for evaluation.
Following the annotation procedure of SIMART~\citep{zhang2026simart}, we use P3-SAM~\citep{ma2025p3sam} to obtain initial segmentations of challenging AI-generated meshes and scans, then manually merge fragments into functional parts and annotate joints.
The source-exclusion comparisons in Appendix~\ref{sec:source-sensitivity} test whether the reconstruction advantage depends on USDCraft-generated references or on agent-generated references in general.

\subsection{Object-category coverage}
\label{sec:benchmark-categories}
Table~\ref{tab:benchmark-categories} lists the object categories in both benchmarks.
For this inventory, USDCraft-bench objects are grouped by function: cabinet variants share one category, as do different microwave designs and storage boxes or cases.
Lightwheel retains its source category names, with equivalent names matched across the two benchmarks.
Under this grouping, USDCraft-bench covers 48 categories and Lightwheel covers 14; seven occur in both, giving $48+14-7=55$ distinct categories in total.

\begin{table}[!htbp]
\centering
\caption{Object categories in USDCraft-bench and Lightwheel.}
\label{tab:benchmark-categories}
\vspace{5pt}
\small
\begin{tabular*}{\linewidth}{@{\extracolsep{\fill}}lll@{}}
\toprule
\multicolumn{3}{@{}l}{\textbf{USDCraft-bench}} \\
\midrule
Chair & Laundry hamper & Stove \\
Clothes dryer & Manipulation assembly & Task lamp \\
Control knob & Microwave & Teapot \\
Dishwasher & Oven & Telescope \\
Door & Paper cutter & Toaster \\
Drawer & Recycling bin & Toaster oven \\
Engine bay & Refrigerator & Toilet \\
Excavator & Rice cooker & Toolbox \\
Fan & Robotic wrist & Utility knife \\
Fire extinguisher & Saucepan & Vanity table \\
Forklift & Service cart & Waffle maker \\
Freezer & Shelving unit & Washing machine \\
Game console & Shower enclosure & Waterwheel \\
Gate & Soap dispenser & Wheelbarrow \\
Hand truck & Storage box/case & Window \\
Laptop & Storage cabinet & Window shutter \\
\midrule
\multicolumn{3}{@{}l}{\textbf{Lightwheel}} \\
\midrule
Blender & Oven & Stove \\
Coffee machine & Range hood & Stovetop \\
Dishwasher & Refrigerator & Toaster \\
Electric kettle & Sink & Toaster oven \\
Microwave & Stand mixer &  \\
\bottomrule
\end{tabular*}
\end{table}

\section{Evaluation Protocols}
\label{sec:appendix-protocols}
\label{sec:evaluation-details}

\subsection{Normalization and geometric evaluation}
For USDCraft-bench, each prediction and reference is independently centered at its static bounding-box center and uniformly scaled to unit longest side.
This transform also applies to joint anchors and prismatic ranges and remains fixed across articulated states; no rotation alignment or per-part fitting is used.
It removes only each object's global translation and scale: part sizes and positions relative to the object, joint anchors, and prismatic ranges are still compared.
This keeps methods without a metric frame, such as image-only generation, from being penalized for global scale.
External predictions are first restored through their recorded input-to-world transforms; USD evaluation merges fixed joints while retaining movable parts.
For Lightwheel, we use the same setting as Instruct-Particulate~\citep{li2026instructparticulate}.
Geometric part correspondence uses Hungarian assignment of centroids. Motion metrics report the fully articulated endpoint.
Chamfer distances sum the two directional means of squared nearest-neighbor distances, with missing-part penalties retained for PC.
We sample 100,000 surface points per asset.

Table~\ref{tab:bench-shared-frame} in Appendix~\ref{sec:reference-frame} instead evaluates every prediction in the reference frame, applying the reference's center and scale to both prediction and reference, so that errors in absolute size and position are retained.
USDCraft's scores change little and remain the best on all nine metrics, whereas image-only Articraft drops sharply because its output lacks a metric scale, and Mini Workflow, which uses the same backbones without source geometry analysis, also drops noticeably; mesh-segmentation baselines are nearly unchanged because they label the input mesh.

\subsection{Part matching, aggregation, and failed outputs}
On USDCraft-bench, a correct movable-part match requires the same joint type and either static gIoU $\geq 0.25$ or centroid separation $\leq 0.600$ times the smaller part's longest side with a longest-side ratio $\leq 2$.
Axis-bearing matches additionally require folded angular error $\leq 15^\circ$; revolute axis-line separation must be $\leq 0.100$ in normalized coordinates.
Explicitly represented unconstrained parts are included in part recovery.
F1 is computed per object from matched-part precision and recall, with zero assigned when both are zero, then averaged over applicable objects.
AE and LE first average over reference joints within each object, assigning penalties of $90^\circ$ and 0.5, respectively, to missing or incorrectly typed matches.
Objects without applicable reference joints are excluded from the corresponding averages.

ArtLLM's metrics exclude 28 cases with no reconstruction due to P3-SAM segmentation failure and use the same 72 successful cases throughout: 45 from the agent-generated subset and 27 from the mixed-source extension.
As a check on case selection, recomputing all methods on these same successful cases gives F1 scores of 19.337\% for ArtLLM, 42.041\% for SIMART, 36.556\% for Particulate, 80.824\% for USDCraft--Sol, and 86.524\% for USDCraft--Astra.

\subsection{Generation scoring}
\label{sec:generation-evaluation-prompt}
Each artifact is judged in a fresh context with a frozen prompt and an evidence package containing the request, part/operation requirements, geometry and joint records, appearance and neutral-material views, and per-joint motion sheets; image-conditioned requests also include the reference image.
Following the LLM-as-a-judge setup~\citep{zheng2023llmjudge}, the judge assigns integer scores from 1 to 5 with evidence-based reasons using the rubric in Appendix~\ref{sec:judge-prompt}.

\section{Additional Results}
\label{sec:additional-experiments}
\subsection{Run-to-run consistency}
\label{sec:harness-repeatability}
To assess run-to-run consistency, Mini Workflow--Astra and USDCraft--Astra are each repeated three times on the same benchmark objects; the other configurations in Table~\ref{tab:harness100} use one run.
Table~\ref{tab:harness-repeatability} reports the means and standard deviations of all nine metrics across the three runs.
USDCraft--Astra varies less than Mini Workflow--Astra on all geometric metrics except OC, where both are equal, and its margin over Mini Workflow--Astra exceeds the run-to-run variation on every metric.

\begin{table}[!htbp]
\centering
\caption{Reconstruction consistency over three Astra runs on USDCraft-bench. Entries report mean $\pm$ standard deviation across runs.}
\label{tab:harness-repeatability}
\vspace{5pt}
\small\setlength{\tabcolsep}{8pt}
\begin{tabular}{llcc}
\toprule
 & Metric & Mini Workflow--Astra & \cellcolor{oursgray}USDCraft--Astra \\
\midrule
Part match & F1(\%)$\uparrow$ & $74.48\pm1.00$ & \cellcolor{oursgray}$84.83\pm0.88$ \\
\midrule
Static geometry & gIoU$\uparrow$ & $0.512\pm0.023$ & \cellcolor{oursgray}$0.694\pm0.010$ \\
 & PC$\downarrow$ & $0.070\pm0.005$ & \cellcolor{oursgray}$0.052\pm0.002$ \\
 & mIoU$\uparrow$ & $0.646\pm0.013$ & \cellcolor{oursgray}$0.784\pm0.009$ \\
\midrule
Articulated geometry & gIoU$\uparrow$ & $0.495\pm0.022$ & \cellcolor{oursgray}$0.670\pm0.009$ \\
 & PC$\downarrow$ & $0.119\pm0.008$ & \cellcolor{oursgray}$0.093\pm0.001$ \\
 & OC$\downarrow$ & $0.018\pm0.001$ & \cellcolor{oursgray}$0.010\pm0.001$ \\
\midrule
Joint axes & AE$^\circ\downarrow$ & $21.14\pm0.21$ & \cellcolor{oursgray}$6.83\pm0.87$ \\
 & LE$\downarrow$ & $0.117\pm0.003$ & \cellcolor{oursgray}$0.044\pm0.004$ \\
\bottomrule
\end{tabular}
\end{table}

\clearpage
\subsection{Additional reconstruction comparisons}
\label{sec:additional-comparisons}
Figure~\ref{fig:supplementary-comparisons} compares six selected objects with GT, Particulate, SIMART, Mini Workflow--Astra, and USDCraft--Astra.
Astra results show the first of three runs.

\medskip
\noindent\begin{minipage}{\linewidth}
    \centering
    \includegraphics[width=\linewidth]{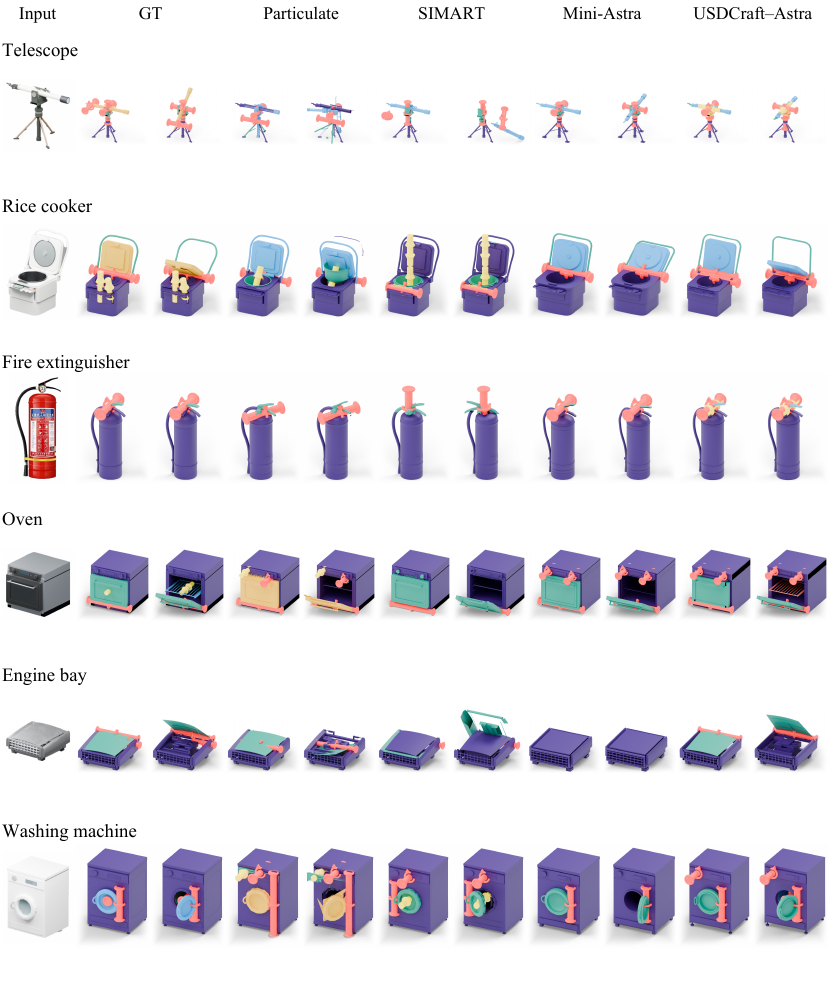}
    \teasercaption{Additional reconstruction comparisons. Each method shows rest and articulated poses from left to right; colors distinguish parts and joint markers follow Figure~\ref{fig:bench-comparison}.}
    \label{fig:supplementary-comparisons}
\end{minipage}

\clearpage
\subsection{Reconstruction galleries}
\label{sec:reconstruction-galleries}
Figures~\ref{fig:gallery-bench} and~\ref{fig:gallery-lightwheel} show high-scoring USDCraft--Astra reconstructions, selected by F1 and static and articulated gIoU across object categories.
Failure cases appear in Appendix~\ref{sec:qualitative-protocol}.

\medskip
\noindent\begin{minipage}{\linewidth}
    \centering
    \includegraphics[width=\linewidth,height=0.8\textheight,keepaspectratio]{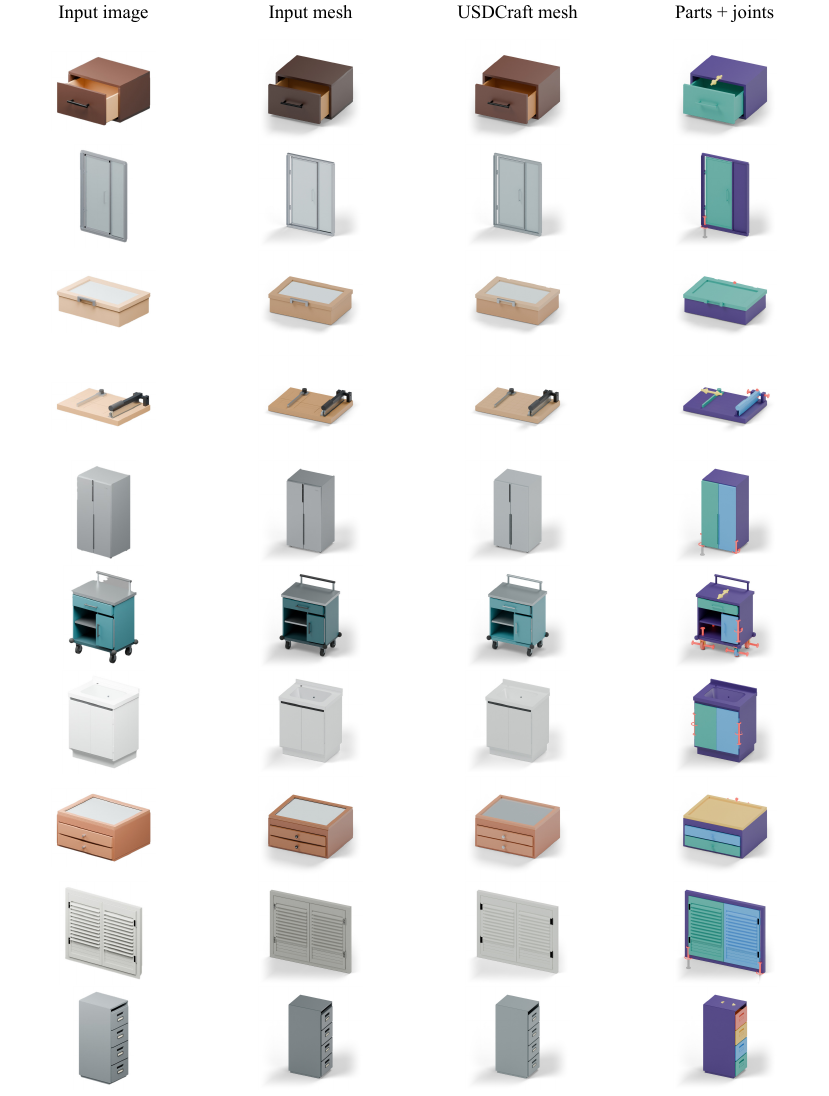}
    \teasercaption{USDCraft-bench reconstruction gallery. Each row shows the input image, source mesh, reconstructed appearance, and part colors with joint markers. Source and reconstruction share a camera and scale.}
    \label{fig:gallery-bench}
\end{minipage}

\clearpage
\noindent\begin{minipage}{\linewidth}
    \centering
    \includegraphics[width=\linewidth,height=0.9\textheight,keepaspectratio]{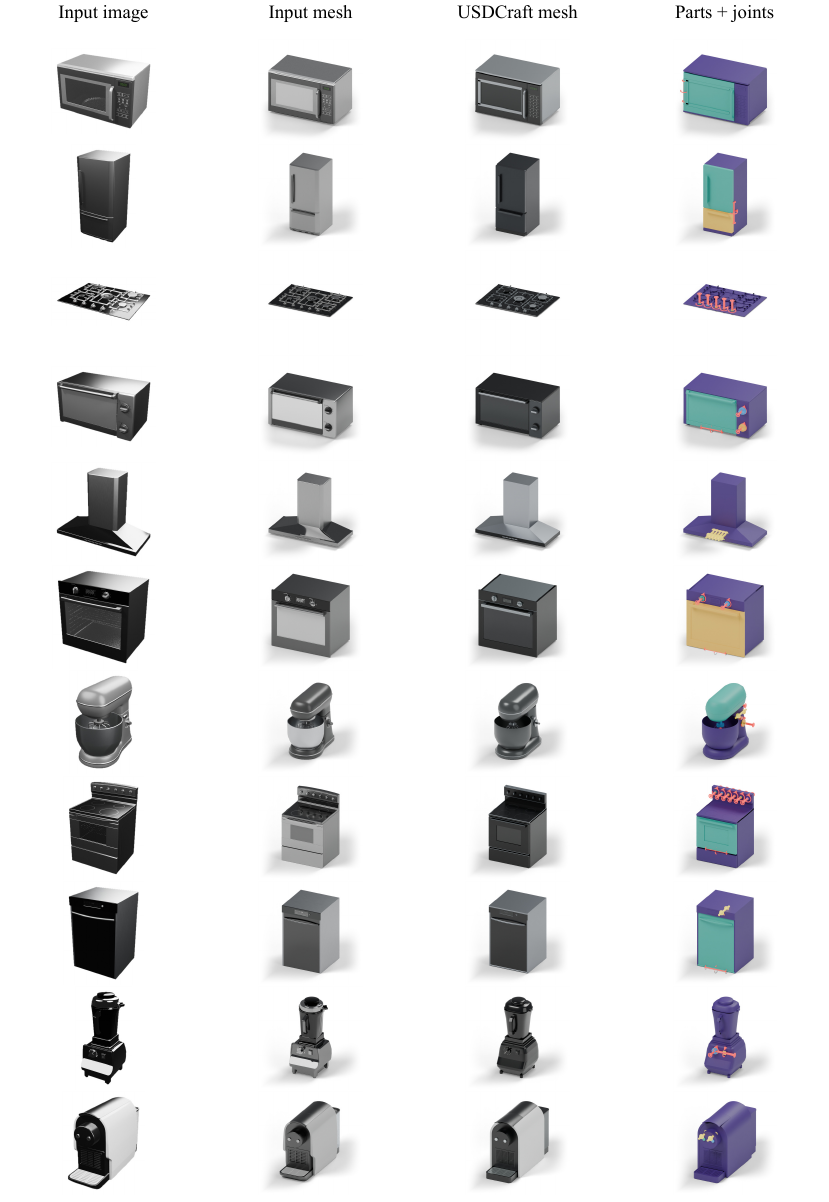}
    \teasercaption{Lightwheel reconstruction gallery using USDCraft--Astra without +G. Columns follow Figure~\ref{fig:gallery-bench}.}
    \label{fig:gallery-lightwheel}
\end{minipage}

\clearpage
\subsection{Generation gallery}
\label{sec:generation-gallery}
Figure~\ref{fig:generation-gallery} presents ten text-conditioned and ten image-conditioned outputs generated with USDCraft--Astra, including complex mechanisms and household objects.
Text panels show object names; representative recorded requests and the shared instructions are given in Appendix~\ref{sec:gallery-prompts}, and image panels show the supplied inputs.

\medskip
\noindent\begin{minipage}{\linewidth}
    \centering
    \includegraphics[width=\linewidth,height=0.8\textheight,keepaspectratio]{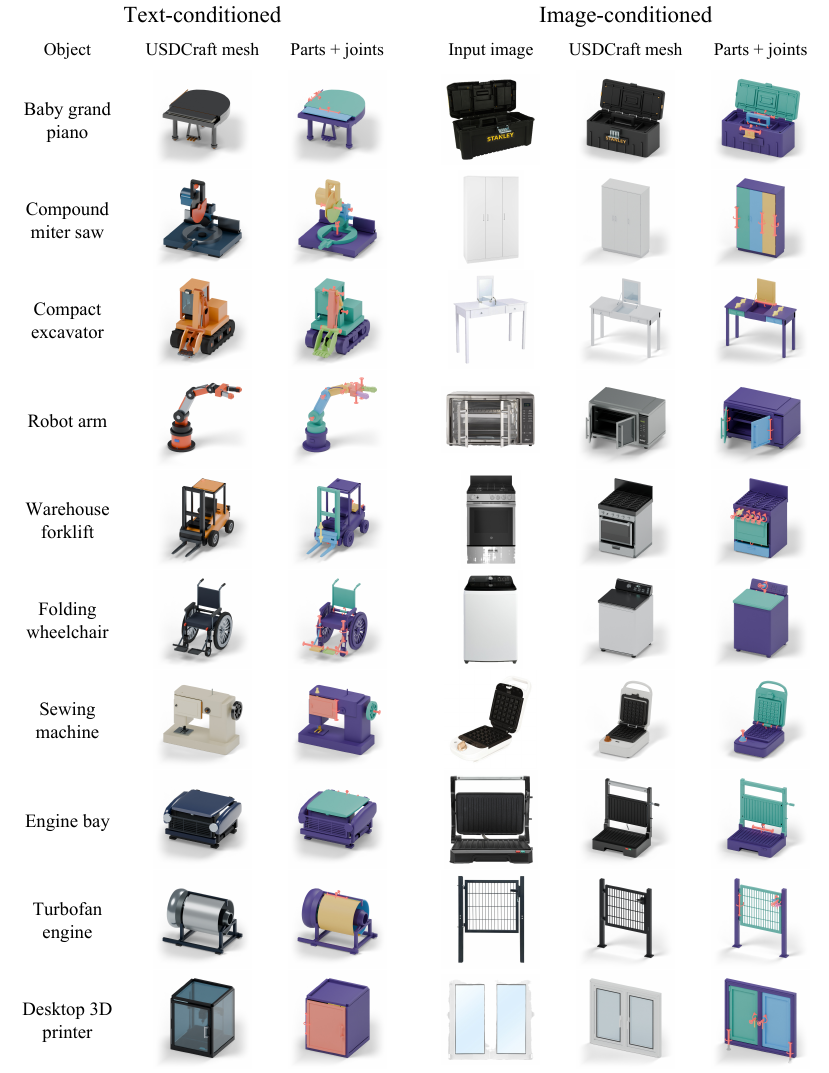}
    \teasercaption{Text- and image-conditioned asset generation. Each example shows its input evidence, the generated mesh with authored appearance, and the same geometry colored by rigid part with joint markers.}
    \label{fig:generation-gallery}
\end{minipage}

\clearpage
\subsection{Sensitivity to benchmark sources}
\label{sec:source-sensitivity}
To test whether results depend on reference provenance, Table~\ref{tab:source-subsets} reaggregates recorded predictions after two source exclusions.
The two settings exclude all USDCraft-generated references or all agent-generated references (USDCraft, Articraft, and Procedura), respectively.
In both settings, USDCraft (Astra or Sol) gives the best result on each metric.

\begin{table}[!htb]
\centering
\caption{Reconstruction after source exclusions. Each panel uses the same objects across methods; USDCraft--Astra and Mini Workflow--Astra average three runs. \textbf{Best}/\underline{second best}. ArtLLM$^{\dagger}$: only success.}
\label{tab:source-subsets}
\vspace{5pt}
\footnotesize\setlength{\tabcolsep}{3.4pt}
\sbox{\papertablebox}{\begin{tabular}{lccccccccc}
\toprule
 & Part match & \multicolumn{3}{c}{Static geometry} & \multicolumn{3}{c}{Articulated geometry} & \multicolumn{2}{c}{Joint axes} \\
\cmidrule(lr){2-2}\cmidrule(lr){3-5}\cmidrule(lr){6-8}\cmidrule(lr){9-10}
Method & F1(\%)$\uparrow$ & gIoU$\uparrow$ & PC$\downarrow$ & mIoU$\uparrow$ & gIoU$\uparrow$ & PC$\downarrow$ & OC$\downarrow$ & AE$^\circ\downarrow$ & LE$\downarrow$ \\
\midrule
\multicolumn{10}{l}{\textit{Excluding all USDCraft-generated references}} \\
ArtLLM$^{\dagger}$ & 18.910 & 0.150 & 0.151 & 0.366 & 0.146 & 0.364 & 0.042 & 61.440 & 0.349 \\
SIMART & 31.632 & 0.222 & 0.140 & 0.452 & 0.212 & 0.206 & 0.016 & 49.643 & 0.252 \\
Particulate & 35.134 & 0.251 & 0.133 & 0.460 & 0.243 & 0.234 & 0.022 & 46.240 & 0.229 \\
Articraft--Astra & 61.021 & 0.160 & 0.107 & 0.406 & 0.153 & 0.194 & 0.023 & 16.936 & 0.114 \\
Mini Workflow--Astra & 72.335 & 0.449 & 0.078 & 0.605 & 0.431 & \underline{0.116} & 0.016 & 23.533 & 0.142 \\
\rowcolor{oursgray}
USDCraft--Sol & \underline{78.202} & \underline{0.540} & \underline{0.061} & \underline{0.663} & \underline{0.521} & 0.117 & \underline{0.014} & \underline{9.331} & \underline{0.086} \\
\rowcolor{oursgray}
USDCraft--Astra & \textbf{83.854} & \textbf{0.636} & \textbf{0.058} & \textbf{0.742} & \textbf{0.618} & \textbf{0.106} & \textbf{0.013} & \textbf{8.507} & \textbf{0.062} \\
\midrule
\multicolumn{10}{l}{\textit{Excluding all agent-generated references}} \\
ArtLLM$^{\dagger}$ & 32.444 & 0.177 & 0.156 & 0.430 & 0.172 & 0.315 & 0.043 & 56.700 & 0.252 \\
SIMART & 27.950 & 0.076 & 0.147 & 0.342 & 0.067 & 0.180 & 0.015 & 58.794 & 0.270 \\
Particulate & 19.375 & 0.111 & 0.172 & 0.397 & 0.100 & 0.264 & 0.024 & 60.983 & 0.255 \\
Articraft--Astra & 63.626 & 0.102 & 0.100 & 0.384 & 0.092 & 0.143 & 0.020 & 8.782 & 0.116 \\
Mini Workflow--Astra & 73.957 & 0.282 & 0.083 & 0.483 & 0.270 & 0.119 & 0.023 & 16.564 & 0.139 \\
\rowcolor{oursgray}
USDCraft--Sol & \underline{77.706} & \underline{0.390} & \underline{0.058} & \underline{0.538} & \underline{0.380} & \underline{0.091} & \underline{0.013} & \textbf{3.341} & \textbf{0.077} \\
\rowcolor{oursgray}
USDCraft--Astra & \textbf{89.827} & \textbf{0.537} & \textbf{0.055} & \textbf{0.659} & \textbf{0.523} & \textbf{0.065} & \textbf{0.003} & \underline{7.535} & \underline{0.084} \\
\bottomrule
\end{tabular}}
\typeout{TABLE-WIDTH-source-subsets=\the\wd\papertablebox; LIMIT=\the\linewidth}
\usebox{\papertablebox}
\end{table}

\subsection{Evaluation in the reference frame}
\label{sec:reference-frame}
Table~\ref{tab:bench-shared-frame} reports USDCraft-bench results without independent normalization of predictions (Appendix~\ref{sec:appendix-protocols}).

\begin{table}[!htbp]
\centering
\caption{USDCraft-bench results evaluated in the reference frame. \textbf{Best}/\underline{second best}. ArtLLM$^{\dagger}$: only success.}
\label{tab:bench-shared-frame}
\vspace{5pt}
\small\setlength{\tabcolsep}{5pt}
\sbox{\papertablebox}{\begin{tabular}{lccccccccc}
\toprule
 & Part match & \multicolumn{3}{c}{Static geometry} & \multicolumn{3}{c}{Articulated geometry} & \multicolumn{2}{c}{Joint axes} \\
\cmidrule(lr){2-2}\cmidrule(lr){3-5}\cmidrule(lr){6-8}\cmidrule(lr){9-10}
Method & F1(\%)$\uparrow$ & gIoU$\uparrow$ & PC$\downarrow$ & mIoU$\uparrow$ & gIoU$\uparrow$ & PC$\downarrow$ & OC$\downarrow$ & AE$^\circ\downarrow$ & LE$\downarrow$ \\
\midrule
ArtLLM$^{\dagger}$ & 19.337 & 0.194 & 0.137 & 0.393 & 0.189 & 0.372 & 0.044 & 61.156 & 0.366 \\
SIMART & 35.957 & 0.282 & 0.124 & 0.483 & 0.272 & 0.216 & 0.020 & 46.656 & 0.262 \\
Particulate & 33.366 & 0.302 & 0.127 & 0.500 & 0.291 & 0.212 & 0.021 & 48.280 & 0.244 \\
Articraft--Astra & 37.398 & -0.054 & 0.151 & 0.261 & -0.058 & 0.231 & 0.052 & 17.900 & 0.163 \\
\midrule
Mini Workflow--Sol & 63.305 & 0.317 & 0.108 & 0.507 & 0.306 & 0.146 & 0.028 & 25.295 & 0.158 \\
Mini Workflow--Astra & 65.970 & 0.378 & 0.098 & 0.563 & 0.364 & 0.145 & 0.030 & 22.504 & 0.140 \\
\midrule
\rowcolor{oursgray}
USDCraft--Sol & \underline{77.541} & \underline{0.551} & \underline{0.062} & \underline{0.674} & \underline{0.532} & \underline{0.117} & \underline{0.014} & \underline{9.155} & \underline{0.071} \\
\rowcolor{oursgray}
USDCraft--Astra & \textbf{83.303} & \textbf{0.667} & \textbf{0.051} & \textbf{0.762} & \textbf{0.644} & \textbf{0.093} & \textbf{0.010} & \textbf{6.833} & \textbf{0.047} \\
\bottomrule
\end{tabular}}
\typeout{TABLE-WIDTH-bench-shared-frame=\the\wd\papertablebox; LIMIT=\the\linewidth}
\usebox{\papertablebox}
\end{table}

\subsection{Token usage and cost}
\label{sec:token-cost}
Table~\ref{tab:token-cost} reports the mean modeling-agent token usage and estimated API cost per asset for the configurations in Table~\ref{tab:harness100}.
With the same backbone, USDCraft uses more input tokens than Mini Workflow at a similar number of turns (Table~\ref{tab:harness100}), since geometric evidence is returned to the agent as text.

\begin{table}[!htbp]
\centering
\caption{Mean modeling-agent token usage (thousands) and estimated cost (USD) for the methods in Table~\ref{tab:harness100}. Cache reads are included in input.}
\label{tab:token-cost}
\vspace{5pt}
\small\setlength{\tabcolsep}{9pt}
\begin{tabular}{lcccc}
\toprule
Method & Input & Cache read & Output & Cost (\$) \\
\midrule
Mini Workflow--Sol & 513.4 & 402.3 & 18.9 & 0.6878 \\
Mini Workflow--Astra & 128.3 & 101.0 & 5.7 & 0.4619 \\
\midrule
\rowcolor{oursgray}
USDCraft--Sol & 1,164.1 & 1,000.3 & 16.5 & 0.9704 \\
\rowcolor{oursgray}
USDCraft--Astra & 560.0 & 508.5 & 3.8 & 0.8510 \\
\midrule
\rowcolor{oursgray}
USDCraft--Claude & 6,493.0 & 6,313.2 & 101.1 & 4.5115 \\
\rowcolor{oursgray}
USDCraft--Gemini & 4,916.8 & 4,250.4 & 80.5 & 0.8643 \\
\bottomrule
\end{tabular}
\end{table}

\section{Real-to-Sim-to-Real Manipulation}
\label{sec:sim2real-protocol}

The manipulation scenes use three real objects: a drawer, a toaster, and a box.
Each object is scanned with LiDAR and RGB on an iPhone Pro, taking approximately two minutes per object.
USDCraft--Astra reconstructs all three objects from their scanned meshes and reference images, including the objects used as supports; all methods use these USDCraft--Astra supports in their scenes.
Figure~\ref{fig:sim2real-asset-preparation} shows the inputs and reconstructed assets.

For the manipulation assets, the modeling agent authors appearance directly in the reconstruction programs: material parameters reproduce the drawer and toaster finishes, while procedural decals represent the box's printed bands and markings, without reusing the scanned textures.

\medskip
\noindent\begin{minipage}{\linewidth}
    \centering
    \includegraphics[width=\linewidth]{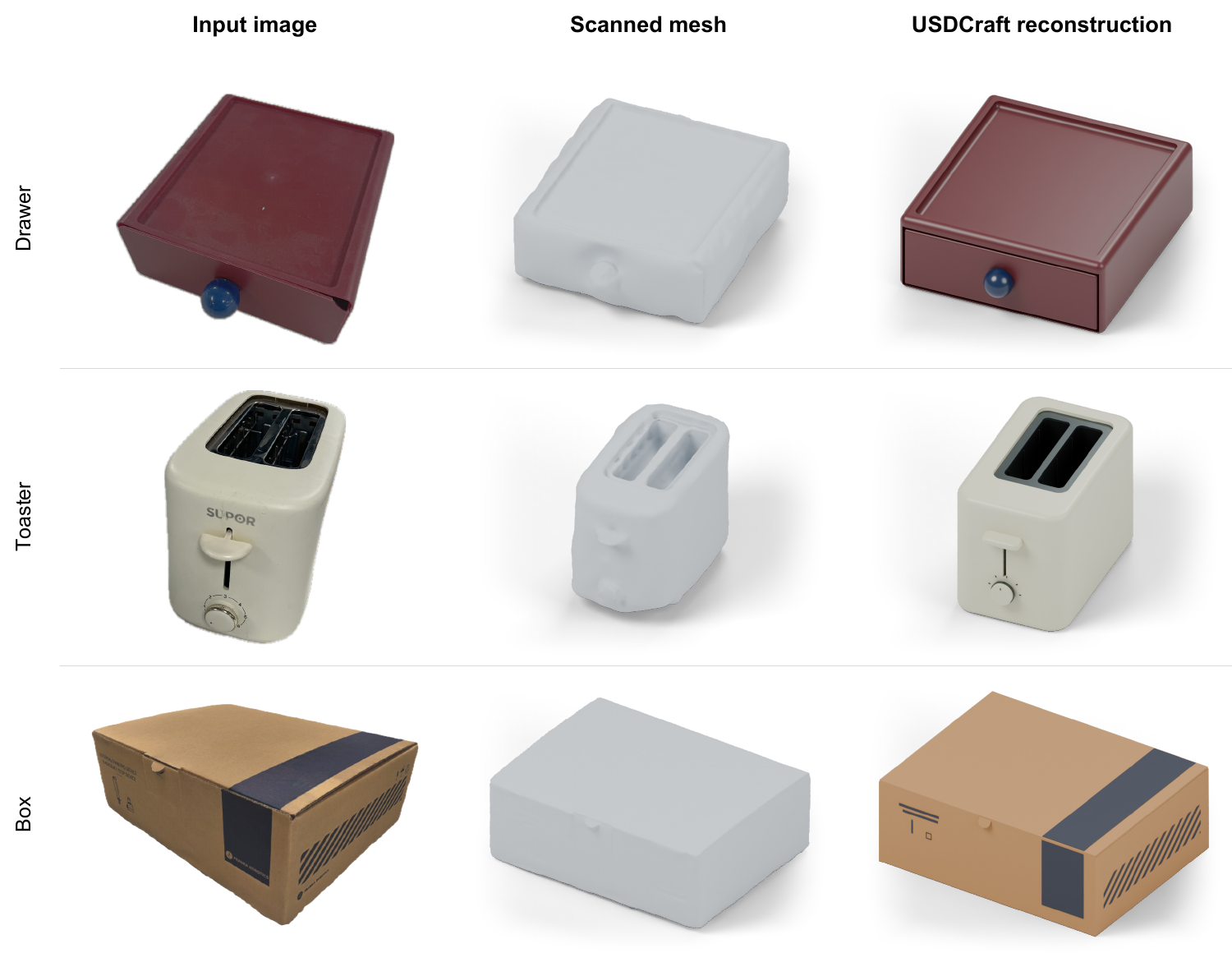}
    \teasercaption{Asset preparation for robot manipulation: reference images, scanned geometry, and USDCraft reconstructions. Scan and reconstruction share the same camera and scale within each row.}
    \label{fig:sim2real-asset-preparation}
\end{minipage}
\medskip

\begin{figure}[!t]
    \centering
    \includegraphics[width=\linewidth]{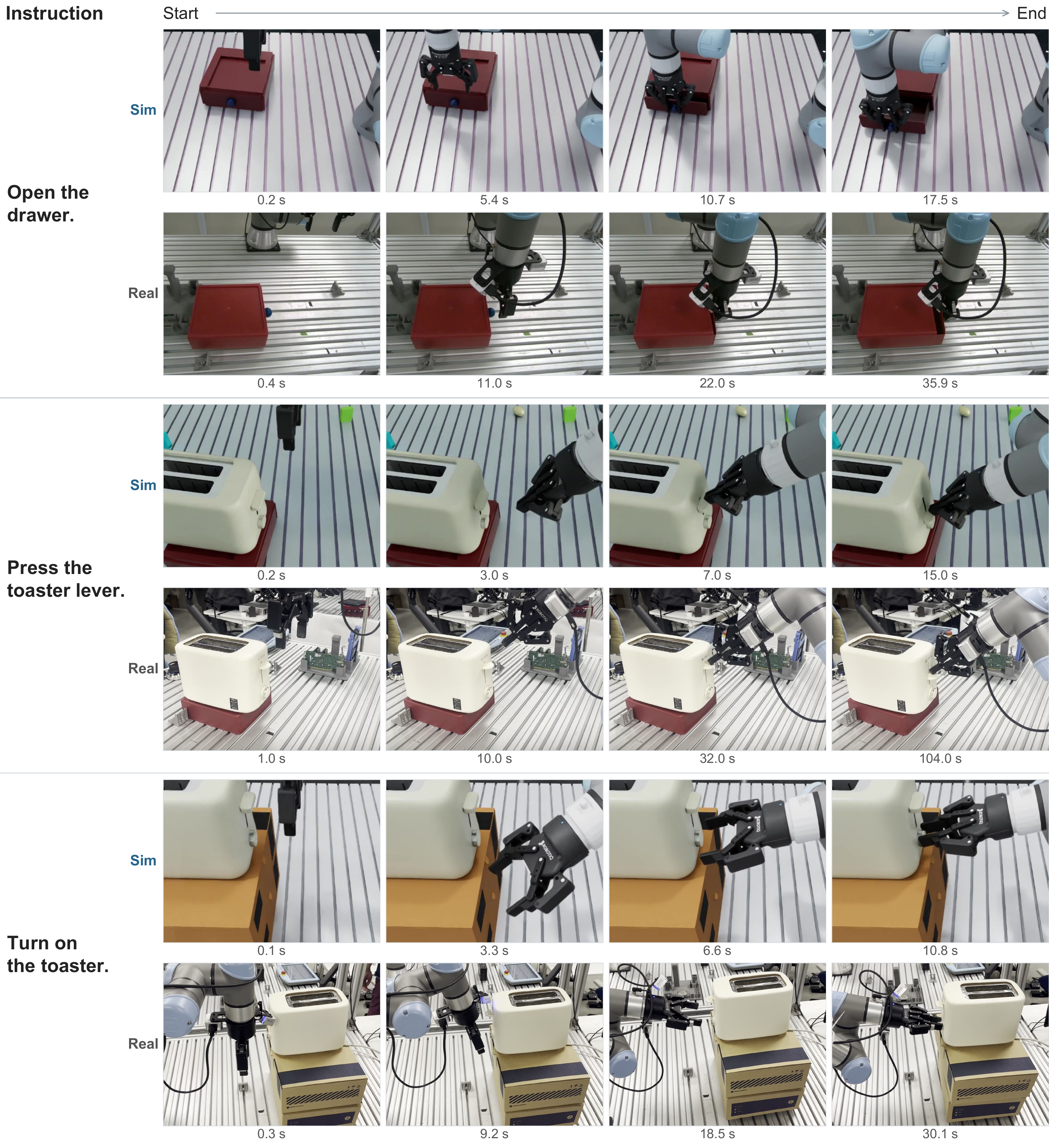}
    \caption{Simulation and real-world execution of the three manipulation tasks. Each row shows four chronological video frames; timestamps refer to its own recording.}
    \label{fig:sim2real-task-sequences}
\end{figure}

To place the toaster within the robot's reach, it is supported by the drawer for lever pressing and by the box for switching.
The corresponding USDCraft assets are stacked in simulation to reproduce these arrangements.
These multi-object scenes make dimension preservation important beyond the manipulated object itself: the support height also determines the toaster's lever and switch positions relative to the robot.
Figure~\ref{fig:sim2real-task-sequences} illustrates the three tasks in simulation and on the real objects.

\paragraph{Robot and policy training.}
We use a UR7e manipulator with a Robotiq 2F-85 gripper and a single external camera.
The policy observes RGB images and outputs absolute end-effector poses.
The robot, table, and camera are calibrated to align the simulated and real workspaces.
For each method--task setting that supports the required interaction, NVIDIA cuMotion automatically collects 200 demonstrations entirely in simulation.
A separate Diffusion Policy is trained once per setting for 80 epochs with batch size 128 and deployed directly on the real robot, without real-world training data or fine-tuning.

\paragraph{Evaluation protocol.}
Each trained policy is evaluated in 20 simulation trials and 20 real-world trials, with randomized object positions and initial robot configurations.
Success requires opening the drawer by at least 5\,cm, pressing the toaster lever down by at least 4\,cm, or rotating the toaster switch by more than $5^\circ$, respectively.
Rollouts stopped because of dangerous contact count as failures; this rule applies to the executed Mini Workflow--Astra drawer rollouts reported as 0\% real-world success.

\paragraph{Baseline asset preparation.}
USDCraft and Mini Workflow receive image and mesh evidence; Particulate receives the mesh, while Articraft receives an image and its generated asset is scaled using the real-scan dimensions.
Particulate recovers the toaster-lever joint but cannot support drawer opening or toaster switching.
No demonstrations are collected for these two settings, and their entries in Table~\ref{tab:sim2real} are recorded as zero success due to asset reconstruction failure.
Its reconstructed bodies cannot stand stably, so they are fixed in Isaac Sim; the reported 50\% simulation success for lever pressing uses this fixed-base condition.

\section{Difficult Geometry and Unsupported Dynamics}
\label{sec:qualitative-protocol}

\begin{figure}[htbp]
    \centering
    \includegraphics[width=\linewidth]{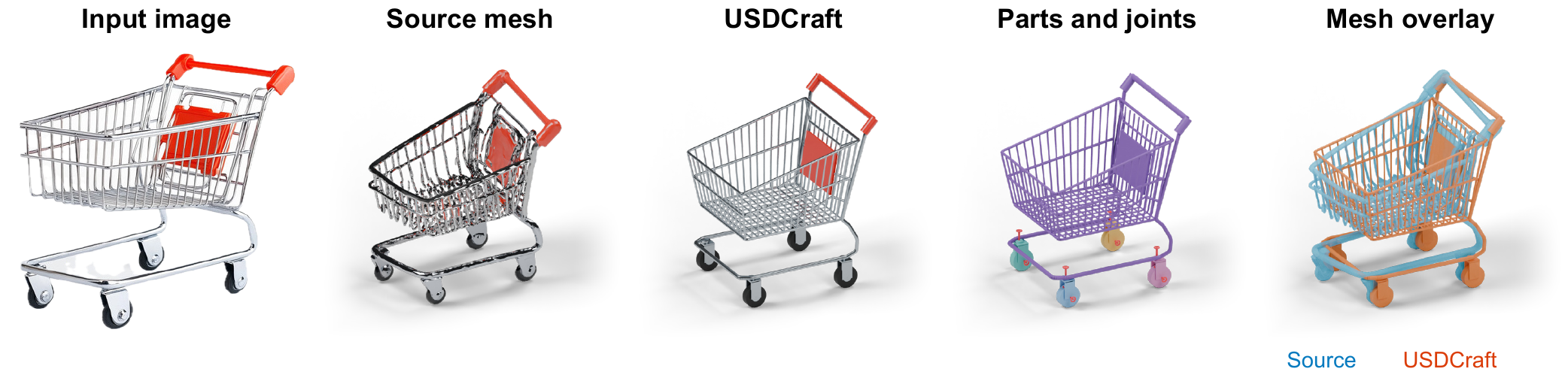}
    \caption{Fine-lattice reconstruction: the shopping cart retains its overall form but differs in local wire geometry. The overlay compares source and reconstruction in their shared coordinate frame.}
    \label{fig:cart-failure}
\end{figure}

Fine lattice structures, such as racket strings and shopping-cart wire frames, remain difficult to reconstruct faithfully.
Their closely spaced members require accurate local geometry and consistent connectivity. Figure~\ref{fig:cart-failure} shows a shopping cart whose overall form is recovered, while wire spacing and local connections differ from the source.

Figure~\ref{fig:flexible-limitations} shows three generated assets with flexible members: a coiled extension cord, a microwave power cord, and an umbrella canopy.
USDCraft can annotate these members and construct their rest geometry, but this does not establish how they bend, stretch, or respond to contact.
For the umbrella, rigid joints on the ribs and stretchers do not model canopy tension or the rigid--flexible coupling, which we have not validated in Isaac Sim.

\medskip
\noindent\begin{minipage}{\linewidth}
    \centering
    \includegraphics[width=0.66\linewidth]{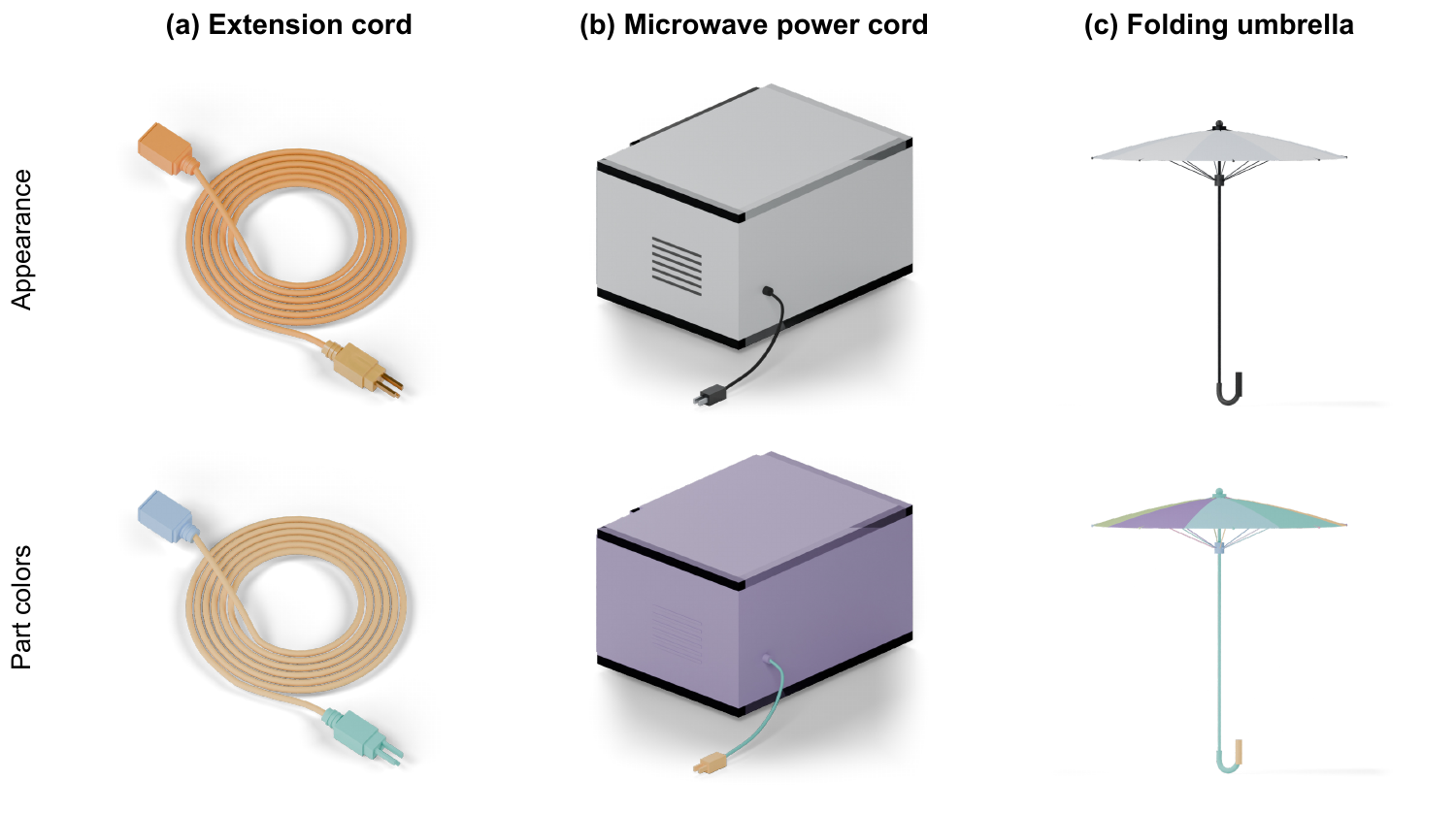}
    \teasercaption{Generated assets with flexible members, shown with native materials and part colors. The umbrella's kinematic display uses annotated attachments.}
    \label{fig:flexible-limitations}
\end{minipage}

\clearpage
\section{Prompts and Authoring Guidance}
\label{sec:prompts}
The instructions for the modeling agent and evaluation rubric are condensed schematics of the recorded prompts; the Mini Workflow prompt is reproduced verbatim.

\subsection{USDCraft reconstruction instructions}
\label{sec:author-prompt}
This schematic preserves the modeling agent role, evidence rules, available tools, and revision loop. SDK references, instance-specific paths, and detailed parameter guidance are omitted.
\begin{lstlisting}[style=paperprompt,basicstyle=\fontsize{7.5}{9}\selectfont\ttfamily]
ROLE AND OUTPUT
You are the reconstruction modeling agent. Reconstruct the supplied object as
an editable, executable model.py with articulated parts and deliberate
physical properties. Own planning, evidence interpretation, construction,
inspection, and repair in one session. The host compiles the program, checks
it mechanically, and persists it.

EVIDENCE AND GEOMETRIC AUTHORITY
Interpret the image and source mesh together. The image establishes identity,
appearance, required visible parts, pose, and normal function. The mesh constrains
reliably observed shape, dimensions, and placement. Preserve the supplied metric
frame, origin, scale, and orientation throughout construction and revision. Axial
geometry encoding is a lossy geometric summary, not a construction blueprint.
Inspect local regions and source views when a summary is ambiguous. Distinguish
measured dimensions from hypotheses; missing surfaces do not establish empty
space. Inspect observed interior geometry before deciding that a compartment is
empty.

FREE AUTHORING AND AVAILABLE TOOLS
Choose the order of construction, inspection, and repair according to the object.
Use the reference image directly; there is no mandatory separate observation
stage. Consult SDK documentation, examples, mesh construction tools, and physical
guidance as needed. Build a coarse executable candidate early, then resolve
consequential uncertainties with focused queries rather than exhaustive
documentation reading.
- measure_mesh / measure_face: inspect source regions and axial detections.
- render_source: request an additional source view to resolve shape or occlusion.
- write_model / apply_patch: create and revise the executable asset program.
- compile_model / validate_model: compile and obtain checks for the current
  revision.
- check_axial / compare_part: compare global or local geometry against the source.
- render_candidate: inspect the compiled candidate at useful views and joint
  poses.

ARTICULATED CONSTRUCTION AND PHYSICS
Use compact, editable geometry that captures characteristic shape and function.
Preserve handles, openings, interiors, contact surfaces, and motion clearances.
Separate genuinely movable assemblies; rigidly attached details share their host.
Choose the fewest degrees of freedom consistent with normal use, including
required secondary adjustments. Do not infer permanent fixation from a stationary
pose. Author masses, inertia, contact materials, joint limits, resistance,
damping, and appropriate drive behavior explicitly. Keep physical assumptions
visible in code. Use a floating standalone root; placement and fixation belong to
the consuming scene.

INSPECTION, RECHECKING, AND COMPLETION
Compile after edits before inspecting or comparing the updated candidate. Review
silhouette, controls, cavities, materials, and functional motion in rendered
views. Use geometric rechecking to locate discrepancies, then inspect the
correspondence and repair supported errors. Preserve reliable dimensions and
required articulation; do not remove necessary motion merely to satisfy a
diagnostic or clearance check. Recheck affected geometry and views after
consequential changes. Report final-revision checks, assumptions, remaining
discrepancies, and unavailable diagnostics. Static validation and posed renders do
not establish successful dynamic simulation.
\end{lstlisting}

\subsection{Mini Workflow reconstruction prompt}
\label{sec:mini-prompt}
Sol and Astra share this prompt. As with USDCraft, which compiles both PhysX and Newton USD entries from the same program, the prompt requests both entries; all experiments in this paper evaluate the PhysX entry only.
\begin{lstlisting}[style=paperprompt,basicstyle=\fontsize{6.5}{7.8}\selectfont\ttfamily,aboveskip=3pt,belowskip=3pt]
Reconstruct the specific object shown in the attached input_image.jpg using input_mesh.glb in the current directory as geometric evidence. Deliver a reusable robotic-manipulation asset with PhysX and Newton/MuJoCo USD entrypoints, generated from one executable modeling definition.

Inspect and measure the scan; preserve reliable dimensions and reconstruct the supported shape. Use the image to interpret identity, appearance, pose, required visible parts and functional structure, including parts missing from the scan. Rebuild all final visual and collision geometry as a compact, editable procedural model in build_asset.py, using measured dimensions, fitted analytic profiles and newly authored topology. The source mesh is reference evidence for measurement and shape interpretation. To produce an independently editable reconstruction, do not copy, instance or reference source vertices/faces, export source segments, or use cleaned, decimated, remeshed or convex-hull versions of the scan as final geometry or colliders. Derive concise shape parameters from measurements instead of reproducing scan samples point by point. Resolve scan noise and fused or missing surfaces through the new construction. Check units and estimate real-world scale if unreliable; state the scale assumption briefly in validation.json. Keep the object's distinctive shape, useful handles, cavities, openings and interior clearances. Check your candidate visually where rendering is available and repair observed defects.

Scale decision for input_mesh.glb: apply scene-node transforms when measuring its axis-aligned bounds. GLB/glTF's format contract is +Y-up and 1 metre per source unit. An owner-declared scale or explicit embedded unit metadata takes precedence; preserve that scale. With format-default metres only, let L be the positive longest scene-bound extent before any discretionary rescaling. Treat the mesh as suspected unit-normalized only if abs(L - t) <= 0.025 * t for t=1.0 or 2.0. Outside those bands, preserve the format metre scale and reliable dimensions; category expectations alone do not authorize resizing. Inside a band, estimate the object's real longest dimension D in metres from the supplied image and request, including attached visible parts; use one isotropic factor D/L. Record L, the unit source, whether normalization was suspected, the chosen factor and (when estimated) D, confidence and a short visual/category rationale in validation.json. This rule supplies no object-specific target dimensions or other model's estimate.

Use only this image and mesh as object evidence. Work independently in this directory using available local tools, without accessing other cases, reference assets, repository code or previous results. Python provides OpenUSD (pxr), NumPy, trimesh and Pillow; use Blender if available. Scripts and intermediate renders may remain here. Put all final USD dependencies inside output/ using relative paths; package the newly constructed geometry without including or referencing the input mesh. Shared geometry/material layers between the two entries are allowed.

Physical requirements:
- Use meters, kilograms and Z-up; author explicit metersPerUnit=1, upAxis=Z and a valid default prim. Create a floating standalone asset, without a world-fixed root, ground plane or PhysicsScene. Configure any articulation consistently with its body/joint tree.
- Give each primary rigid body deliberate positive mass and collision geometry preserving openings and manipulation clearances. Provide consistent center of mass/inertia, or deliberately use collider-derived inertia/COM and record that choice. Every collider must resolve a physics material with explicit contact friction satisfying static >= dynamic >= 0 and nonnegative restitution. Contact friction is distinct from joint resistance; absolute static/dynamic joint-friction efforts must also satisfy static >= dynamic >= 0.
- Use the fewest physical degrees of freedom preserving ordinary use. Doors, lids, drawers and other moving pieces are separate rigid bodies with correct joints, anchors, axes and applicable limits; rigid objects need no invented joints. For every scalar joint, explicitly choose dry resistance, viscous damping, drive/passive behavior, and physical effort/velocity bounds or justified unbounded/not-applicable values. Zero resistance/damping is valid when deliberate. Ordinary passive interaction must not use target-position drives that automatically perform the interaction; a zero-stiffness damping drive is permitted for passive damping on PhysX.
- For a door/lid/slider expected to hold position under gravity, estimate the moving mass and center-of-mass lever arm or axial load, then select an actual hold mechanism with sufficient capacity in the relevant poses. Document the load estimate and hold capacity briefly per applicable joint. Free-fall, contact-supported or unsupported-hold mechanisms must be explicitly identified. Damping alone cannot provide static gravity hold. Do not apply a single unexplained friction value to all joints.
- Sample motion limits and intermediate poses; compute the maximum absolute residual gravity effort after explicit spring/counterbalance/force-drive contributions, including mechanically connected downstream mass and the modeled default payload. Verify static friction covers that residual. Where positive residual is held by dry friction, set Newton uniform DoubleArray mjc:solimpfriction = [0.999, 0.9999, 0.001, 0.5, 2.0] on the MjcJointAPI joint. The first value d0 must be >= 0.999 to reduce soft-constraint creep. This is a pose-hold authoring policy, not runtime proof or PhysX equivalence; zero residual and deliberately free joints need no such tuning.
- Both entrypoints preserve the same geometry, masses, contact materials, joints and limits. Implement backend-supported passive/active behavior and record genuine capability differences; do not claim identical runtime behavior or turn a coefficient into absolute force/torque without a physical derivation.

Compact USD mapping reference (use raw typed attributes/API tokens if optional schema Python modules are absent):
* Common: UsdPhysics.RigidBodyAPI, MassAPI and CollisionAPI; UsdPhysics.MaterialAPI on UsdShade.Material, bound with materialPurpose="physics". Scalar joints use UsdPhysics.RevoluteJoint or PrismaticJoint. USD angular limits/velocities use degrees; scalar effort is N*m for hinges, N for sliders.
* PhysX absolute resistance: apply API token PhysxJointAxisAPI:angular (or :linear); Float attributes physxJointAxis:angular:staticFrictionEffort, :dynamicFrictionEffort, :viscousFrictionCoefficient, :maxJointVelocity (replace angular with linear for sliders). Convert angular viscous coefficient from N*m*s/rad to N*m*s/degree by multiplying pi/180. Alternative load-proportional resistance uses PhysxJointAPI plus Float physxJoint:jointFriction; that dimensionless coefficient is not a hold torque. Passive UsdPhysics.DriveAPI:angular/linear may use stiffness=0, targetVelocity=0 and explicit damping; convert angular damping from per-radian to per-degree. Avoid double-counting viscous resistance.
* Newton passive scalar joints: apply MjcJointAPI; uniform Double mjc:frictionloss is an absolute dry-friction effort and mjc:damping is viscous damping in SI/radian units. For an absolute PhysX static/dynamic model, use static effort as frictionloss, add viscous terms to damping, and note that Newton does not separately reproduce the dynamic dry-friction effort. Load-proportional friction has no verified direct mapping. Use native mjc:stiffness and mjc:springref for real passive springs (radians for angular springref). Deactivate shared DriveAPI on passive Newton joints. On articulation roots use NewtonArticulationRootAPI and Bool newton:selfCollisionEnabled. Active mechanisms need suitable backend actuator handling; state any unimplemented mapping rather than inventing equivalence. Velocity-limit metadata alone does not establish Newton runtime enforcement.

Deliver build_asset.py plus output/physx.usd, output/newton.usd and output/validation.json. Running the build script must regenerate both entries from the same definition and declared local inputs. Verify both entries open independently, have the declared frame/default prim and complete local dependency closure; inspect authored body, material and joint fields. Keep validation.json concise: scale and inertia choices, per-entry static check results, per-joint resistance/damping/drive choices and gravity-hold applicability with estimates, backend degradations, and runtime status for each engine. Record runtime as "not_tested" unless an actual engine test ran; static USD checks do not prove simulation stability or motion correctness. Fix discovered errors. Final response: only the three output paths and build_asset.py, or FAILED if required outputs cannot be produced.
\end{lstlisting}

\subsection{Generation-evaluation rubric}
\label{sec:judge-prompt}
This schematic summarizes evidence review, the five scoring dimensions, score anchors, and response fields. The full prompt additionally expands the score anchors for each dimension and specifies the complete JSON schema.
\begin{lstlisting}[style=paperprompt,basicstyle=\fontsize{7.5}{9}\selectfont\ttfamily]
ROLE AND EVIDENCE BOUNDARY
Evaluate one anonymous generated asset against its request, required parts and
operations, and reference image when supplied. Use only the given evidence
package. Do not infer the method, inspect other assets, repair the output, or rank
methods.

REQUIRED REVIEW
Read the task, manifest, geometry, and joint records. Inspect every supplied
reference, appearance, structure, and motion image, including all views and
sampled joint states. Check every required part and operation. For each joint,
review type, parent and child, axis, anchor, limits, and its relation to the
observed motion. Cite specific files, views, or fields for each conclusion.
Distinguish missing evidence from visible defects.

FIVE SCORING DIMENSIONS
1. Condition adherence: Does the asset match the requested object, key form,
  features, constraints, and applicable reference image? Follow requested changes;
  background, lighting, and camera viewpoint are not required generated content.
2. Structural quality: Are parts credibly constructed and connected, with
  reasonable proportions, thickness, support, layout, and space for required
  motion?
3. Articulation functionality: Does every required operation have suitable
  independent parts, joint types, axes, anchors, limits, and a credible sampled
  motion path? One working joint does not establish that all requested operations
  are supported.
4. Completeness: Are the complete assembly and its necessary static and
  movable components present, without major omissions or truncation?
5. Appearance: Are surfaces, materials, colors, and finish clear,
  coherent, and appropriate? Judge materials from original-material appearance
  views.

SHARED SCORE ANCHORS
1 = Severe failure: the dimension fails fundamentally.
2 = Major problems: recognizable intent, but major requirements fail.
3 = Basically valid: the dimension is established, with conspicuous defects.
4 = Well completed: important requirements are met, with only local minor
problems.
5 = Complete and credible: no obvious defect in the supplied evidence. Score each
dimension independently with a dimension-specific reason. Do not reward extra
decoration or part count, demand photorealism, or calibrate scores to other
assets.

EVIDENCE LIMITS AND EXCEPTIONS
Motion views show static kinematic samples, not physics simulation. They do not
prove stability, load bearing, friction, collision-free dynamics, or simulator
success. Use insufficient_evidence with a null score when missing data or
occlusion prevents judgment. A confirmed asset defect receives a low score, not an
evidence exemption. Use not_applicable only for articulation when the task
requires no movable operation; a missing required joint is a failure. Do not
penalize an upstream authoring status.

RESPONSE
Return one JSON object with, for each dimension, a score (1..5 or null), a status
(scored / insufficient_evidence / not_applicable), a concrete reason, and cited
evidence; also list major defects and evidence gaps.
\end{lstlisting}

\subsection{Lightwheel articulation conventions (+G)}
\label{sec:lightwheel-guidance}
For the +G setting in Table~\ref{tab:lightwheel}, every case receives the complete text below, covering all 14 categories, and the modeling agent decides which descriptions apply.
The text contains no instance-specific dimensions, part masks, or joint annotations.
The descriptions summarize category-level articulation conventions of the Lightwheel assets; the comparison with other methods in Section~\ref{sec:exp-reconstruction} refers to the unguided setting.

\begingroup
\fontsize{8.5}{10}\selectfont
\setlength{\parskip}{3pt}
\setlength{\parindent}{0pt}
\textbf{Category selection.} The following guidance is the same for every input and lists multiple possible object categories. Determine the subject category and relevant mechanisms from the supplied semantic image and cleaned mesh yourself. No category is assigned to this case. Apply only descriptions supported by the input; do not combine unrelated categories or invent parts just because they appear in this reference.\par
\textbf{Shared guidance.} Use the image and cleaned mesh to identify the main body, operating parts, controls and access covers. Inspect seams and interfaces before deciding a part is fixed. Keep integral shells, trim and printed markings fixed; give separate motion only to evidenced functional assemblies. Preserve the measured rest shape and frame, and state uncertain joint ranges as hypotheses. Choose the joint from its mechanical constraint: hinge, rail, or truly unconstrained removal; removability alone does not imply a floating joint. Check each exported joint axis in the frozen world frame, including joint-frame rotations, and inspect a posed render for the intended motion.\par
\textbf{Blender.} Keep the motor housing fixed. Inspect the lift-off jug, removable or hinged lid, independent switches and rotary dials; the handle belongs to the jug. Represent blade rotation only when blades and their shaft are evidenced.\par
\textbf{Coffee machine.} Inspect top and rear access covers, container lids, reservoirs, mechanical buttons and trays separately; use seams and pivots to distinguish hinged covers, sliding parts and lift-off pieces. Keep enclosure and brew-head supports fixed. Do not assume an unseen tank or portafilter, or split a seated tray and grate without an independent removal interface.\par
\textbf{Dishwasher.} Keep the cabinet fixed. Inspect the bottom-hinged door, pull-out racks and evidenced control actuators; the door handle follows the door. Keep decorative panels, seals and stationary plumbing fixed.\par
\textbf{Electric kettle.} For a separate power dock, allow the jug to lift off; keep the jug handle and spout attached. Inspect the lid hinge or lift-off rim and mechanical release/power controls; an integral corded base stays integral.\par
\textbf{Microwave.} Inspect the door hinge or drawer rails, mechanical opening button, knobs and any visible turntable. Keep enclosure and window fixed to their supports; touch icons are not moving buttons. Do not invent hidden turntable geometry.\par
\textbf{Oven.} Inspect the door hinge, sliding racks and mechanical control knobs or buttons. Keep the enclosure fixed and handles/glass attached to their door; decorative seams do not establish extra doors.\par
\textbf{Range hood.} Inspect removable filter panels, any adjustable visor and mechanical controls. Keep the hood, duct and stationary grilles fixed; model fan rotation only if the rotor is evidenced, not by rotating the grille.\par
\textbf{Refrigerator.} Inspect each door and its hinge, pull-out drawers and separately seated trays or shelves. Keep the cabinet fixed and handles attached to doors; fixed trim and sealed panels do not become articulated parts.\par
\textbf{Sink.} Keep basin, deck and mounted faucet base fixed. For a swivel spout, use the bearing collar at the mounting base to determine the pivot and world-space axis; the curved or sloped outlet tube is not the swivel axis. Treat the control lever separately. Give racks, baskets and strainers independent removal only where a separate seating interface is supported; keep integral surfaces fixed.\par
\textbf{Stand mixer.} Inspect the head hinge or bowl-lift mechanism, detachable bowl and tool coupling, and mechanical speed controls. Keep the stand fixed; rotating tool motion belongs at its shaft. Do not add both head-tilt and bowl-lift mechanisms without evidence.\par
\textbf{Stove.} Inspect oven doors, sliding racks or storage drawers, and mechanical burner controls where present. Keep the stove body and burner supports fixed; flames and printed control marks are not articulated bodies.\par
\textbf{Stovetop.} Keep the cooking surface and stationary burner supports fixed. Inspect rotary knobs, mechanical buttons and visibly removable grates or caps; touch markings are static. Do not add an oven or hidden mechanism.\par
\textbf{Toaster.} Inspect the lowering lever and its vertical guide, browning knob, mechanical buttons and removable crumb tray. Keep the shell and slot rims fixed; do not turn slot openings or printed settings into separate moving parts.\par
\textbf{Toaster oven.} Inspect the door hinge, sliding cooking rack or tray, and mechanical knobs/buttons. Keep enclosure fixed and handle/glass attached to the door; keep stationary heating elements fixed.\par
\endgroup

\subsection{Generation gallery prompts}
\label{sec:gallery-prompts}
Table~\ref{tab:gallery-prompts} lists three representative object-specific text requests from Figure~\ref{fig:generation-gallery}; the other seven follow the same template.
Each request is followed by the same shared instructions below; together they reproduce the complete recorded text prompt.

\noindent\textbf{Shared instructions.}
Use realistic scale, plausible materials, solid part thickness and sufficient clearance through the requested motions. Every requested moving component must be a separate articulated rigid part with appropriate joint limits and physical properties. Define the stated starting pose at joint zero. Simplify hidden mechanisms while preserving the named normal-use motions. Include only the complete standalone assembly, without people, surrounding scenery, brand logos or loose props.

\begingroup
\fontsize{8}{9.5}\selectfont
\setlength{\tabcolsep}{4pt}
\renewcommand{\arraystretch}{1.12}
\begin{longtable}{@{}p{.18\linewidth}p{\dimexpr.82\linewidth-8pt\relax}@{}}
\caption{Representative text prompts for the generation gallery; append the shared instructions to each row.}\label{tab:gallery-prompts}\\
\toprule
Object & Object-specific request \\
\midrule
\endfirsthead
\multicolumn{2}{l}{\tablename\ \thetable\ (continued)}\\
\toprule
Object & Object-specific request \\
\midrule
\endhead
\bottomrule
\endfoot
Baby grand piano & Create a baby grand piano with an opening lid and keyboard cover. Include the curved case on three rigid legs, a large top lid hinged along the straight long side, a separate hinged keyboard fallboard and a hinged lid-prop stick mounted to the case. Model a complete recognizable keyboard as fixed geometry, a recessed soundboard and simplified strings. The lid, fallboard and prop are independent limited revolute joints, without a closed-loop support constraint. Start with both covers closed and prop stowed. Approximate length 1.5 m. \\
\addlinespace[5pt]
Compact excavator & Create a compact excavator with a three-stage digging arm. Include a fixed track-shaped undercarriage, a cab and upper platform on one vertical slew joint, a boom on a horizontal shoulder hinge, a stick on a parallel elbow hinge and a hollow toothed bucket on a wrist hinge. Add one side-hinged cab door. Author simplified hydraulic-cylinder appearances without redundant closed-loop constraints. Use limited boom, stick and bucket travel with meaningful folded and reaching poses. Tracks remain rigid decorative assemblies. Approximate chassis length 2.5 m. Start with door closed and arm in a compact collision-free resting pose. \\
\addlinespace[5pt]
Desktop 3D printer & Create an enclosed desktop Cartesian 3D printer. Include a rigid cubic frame, one left-hinged transparent front door, fixed transparent side panels, a print head sliding along an X carriage nested on a vertically sliding Z gantry, and a print bed sliding front to back on Y rails. Provide three independent prismatic axes plus the door hinge. Represent belts and wiring as fixed visual details rather than articulated links. Start with door closed, bed centered and nozzle safely above the bed. Approximate outer width 0.5 m. \\
\addlinespace[5pt]
\end{longtable}
\endgroup

\noindent\textbf{Image-conditioned requests.}
The image-conditioned examples use the supplied reference image and the following shared request:
\begin{quote}
Reconstruct the primary subject object shown in the attached image as a complete, functionally usable, simulator-ready 3D asset.
\end{quote}

\subsection{Source geometry analysis example}
\label{sec:axial-example}
The following excerpt shows the axial encoding of the scanned toaster used in the real-to-sim-to-real experiments.
It retains the recorded metric bounds (rounded to four decimals), three representative XY slices, and one fine-resolution boundary summary; other slices and metadata are omitted for readability.

\begin{lstlisting}[style=paperprompt,basicstyle=\fontsize{7.5}{8.4}\selectfont\ttfamily,breakatwhitespace=false,literate={~}{{$\sim$}}1]
AXIAL ENCODING
frame: world, Z-up, meters; isotropic
bounds_min_xyz: -0.0717, -0.1263, 0.0000
bounds_max_xyz:  0.0717,  0.1264, 0.1682
extent_xyz:      0.1435,  0.2527, 0.1682
cells: . = unknown, ~ = sampled source surface
fine_grid: dims_xyz=112,192,128  pitch=0.0013  grid_min_xyz=-0.0737,-0.1263,-0.0001
slice_stack: axis=Z  dims_xyz=20,32,24  pitch=0.0079  grid_min_xyz=-0.0790,-0.1263,-0.0107
             slice_order=-Z->+Z  rows=+Y->-Y  cols=-X->+X
\end{lstlisting}

\noindent Each column below is one XY slice at an increasing height. Rows run from $+Y$ to $-Y$, and columns from $-X$ to $+X$; $\sim$ denotes sampled surface support and \texttt{.} denotes unknown geometry.
\begin{lstlisting}[style=paperprompt,basicstyle=\fontsize{7.4}{7.6}\selectfont\ttfamily,columns=fixed,literate={~}{{$\sim$}}1]
[PATTERN e]              [PATTERN m]              [PATTERN u]
run e: slices=4:5         run m: slices=12:13       run u: slices=20:21

......~~~...........    ....~~~~~~~~~~~.....    ......~~~~~~~~~.....
....~~~~~~~.........    ...~~.........~~....    ....~~~.......~~....
..~~~~~..~~~~~~~....    ..~~...........~~...    ...~~...........~~..
..~~...........~~~..    ..~.............~~..    ..~~.............~..
..~..............~~.    .~~..............~~.    ..~...~~~~~.~~~..~~.
.~~...............~.    .~~...............~.    ..~..~~..~~~~~~~..~.
.~~...............~.    .~................~.    ..~..~~...~~...~..~.
.~~...............~.    .~................~.    .~~..~~...~~...~..~.
.~~...............~.    .~................~.    .~...~~..~~~...~..~.
.~~...............~.    .~................~.    .~...~~...~~...~..~.
.~................~.    .~................~.    .~~..~~..~~~...~..~.
.~................~.    .~................~.    .~~..~~..~~~...~..~.
.~................~.    .~................~.    ..~..~~..~~~...~..~.
.~................~.    .~................~.    ..~..~~..~~~...~..~.
.~................~.    .~................~.    ..~..~~..~~~...~..~.
.~................~.    .~................~.    ..~..~~..~~~..~~..~.
.~................~.    .~................~.    ..~..~~..~~~..~~..~.
.~................~.    .~................~.    ..~..~~..~~~..~~..~.
.~................~.    .~................~.    ..~..~...~~~..~~..~.
.~................~.    .~................~.    .~~..~...~~~..~~..~.
.~................~.    .~................~.    .~~..~~~~~~~~~~~.~~.
.~................~.    .~................~.    .~..........~~~~.~~.
.~................~.    .~................~.    .~...............~~.
.~...............~~.    .~................~.    .~...............~~.
.~..............~~..    .~...............~~.    .~~..............~~.
.~..............~~..    .~...............~..    ..~..............~..
.~~............~~~..    .~~..............~..    ..~~............~~..
..~~..........~~~~..    ..~~............~~..    ...~~...........~...
...~~~~.....~~~~....    ...~~~.........~~...    ...~~~~.......~~~...
.....~~~~~~~~~......    .....~~~~~~~~~~.....    ......~~~~~~~~~.....
.........~~.........    ....................    ....................
....................    ....................    ....................
\end{lstlisting}

\noindent The upper slice (pattern \texttt{u}) exposes sampled boundaries around the toaster slots, while the lower slices primarily trace the outer shell. Unknown cells do not establish empty space or solid interiors.
The corresponding fine-grid summaries retain local bounds and contours, for example:
\begin{lstlisting}[style=paperprompt,basicstyle=\fontsize{7.5}{8.4}\selectfont\ttfamily,breakatwhitespace=false]
s12:13 mz=64:70 xy=4:109,15:192(f=.03;p=4,43 16,22 28,17 80,15 100,28 106,49 109,153 85,186 38,192 22,187 12,176 5,145)
\end{lstlisting}
\noindent Here, \texttt{mz} and \texttt{xy} give half-open index ranges in the fine grid, \texttt{f} is the fraction of sampled cells within the XY bounding box, and \texttt{p} lists a simplified boundary contour. These measurements let the modeling agent locate and size geometry beyond the coarse character grid.

\end{document}